\documentclass[11pt]{article}
\usepackage{fontspec}
\usepackage{xeCJK}
\setCJKsansfont{HaranoAjiGothic-Regular.otf}
\setCJKmonofont{HaranoAjiGothic-Regular.otf}
\usepackage[margin=1in]{geometry}
\usepackage{graphicx}
\usepackage{booktabs}
\usepackage{tabularx}
\usepackage{array}
\usepackage{amsmath}
\usepackage{enumitem}
\usepackage{hyperref}
\usepackage{url}
\usepackage{multirow}
\usepackage{xcolor}
\usepackage{changepage}
\usepackage{subcaption}
\definecolor{promptgreen}{RGB}{70,120,50}
\usepackage{placeins}
 \usepackage{comment}

\hypersetup{colorlinks=true,linkcolor=blue,citecolor=blue,urlcolor=blue}
\setlist[itemize]{leftmargin=1.5em}

\providecommand{\tightlist}{%
  \setlength{\itemsep}{0pt}\setlength{\parskip}{0pt}}

\newcommand{\hflink}{https://huggingface.co/collections/pfnet/plamo-2-vl}
\newcommand{\huggingfaceicon}{%
  \raisebox{-1.5pt}{\includegraphics[height=1.55em]{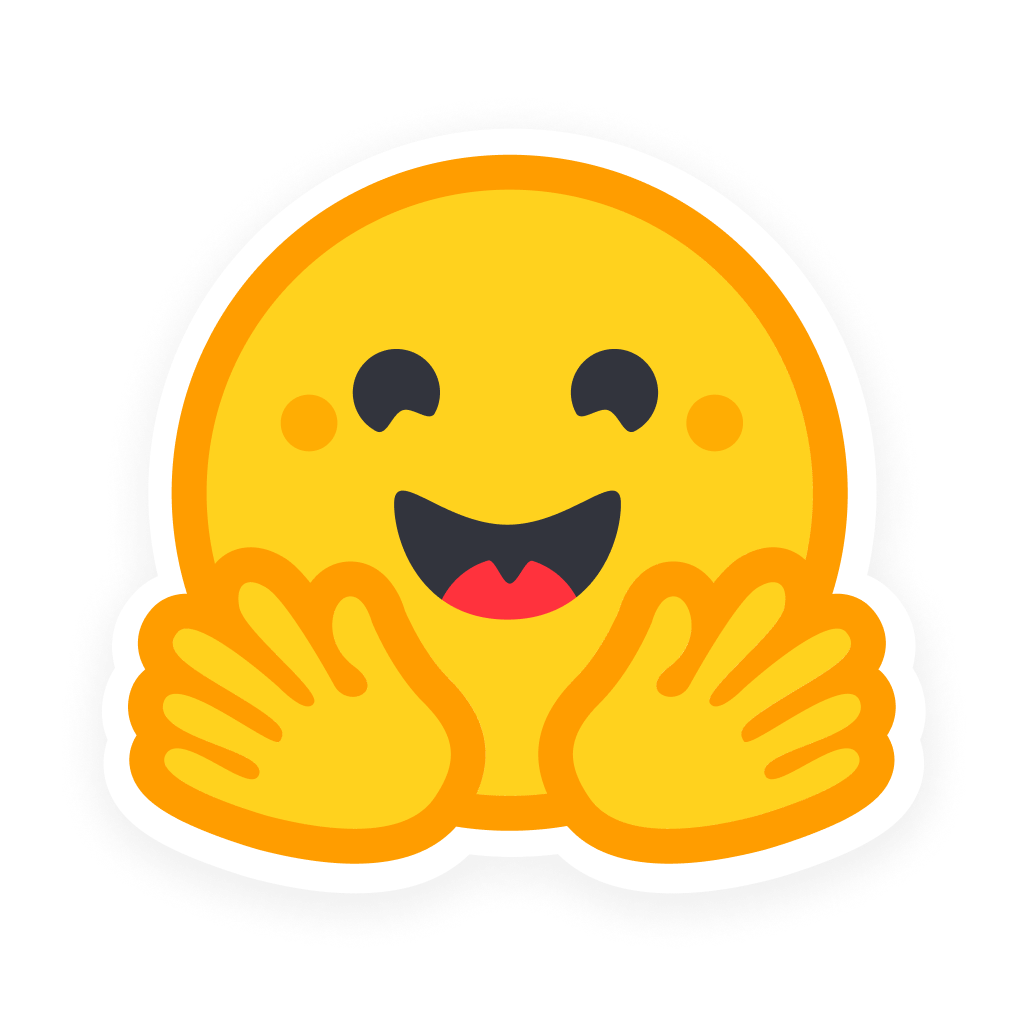}}%
}

\title{PLaMo 2.1-VL Technical Report}
\author{
\normalsize{Preferred Networks, Inc. (PFN)}
}
\date{}

\begin{document}
\maketitle

\vspace{-3.5em}

\begin{center}
    \small
    \begin{tabular}{@{}rl@{}}
    \huggingfaceicon & \url{\hflink}
    \end{tabular}
\end{center}
\vspace{0.5em}

\par\noindent
\makebox[\textwidth][c]{
  \href{\hflink}{\includegraphics[width=0.70\textwidth]{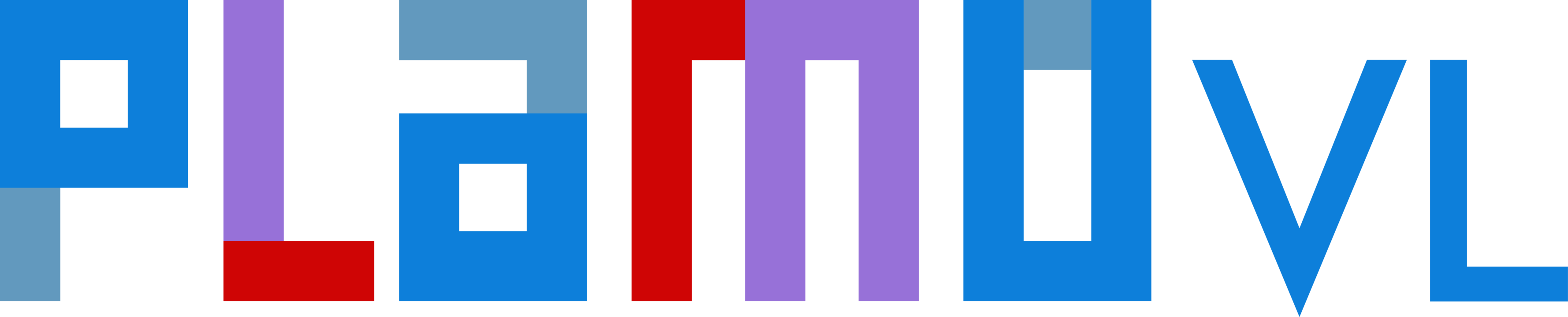}}
}
\par

\begin{abstract}
We introduce PLaMo 2.1-VL, a lightweight Vision Language Model (VLM) for autonomous devices, available in 8B and 2B variants and designed for local and edge deployment with Japanese-language operation.
Focusing on Visual Question Answering (VQA) and Visual Grounding as its core capabilities, we develop and evaluate the models for two real-world application scenarios: factory task analysis via tool recognition, and infrastructure anomaly detection. We also develop a large-scale synthetic data generation pipeline and comprehensive Japanese training and evaluation resources. PLaMo 2.1-VL outperforms comparable open models on Japanese and English benchmarks, achieving 61.5 ROUGE-L on JA-VG-VQA-500 and 85.2\% accuracy on Japanese Ref-L4. For the two application scenarios, it achieves 53.9\% zero-shot accuracy on factory task analysis, and fine-tuning on power plant data improves anomaly detection bbox + label F1-score from 39.7 to 64.9.
\end{abstract}
\section{Introduction}
\label{sec:introduction}
In recent years, autonomous devices such as drones, robots, automobiles, and surveillance cameras have emerged as important application targets for large language models and generative AI. In these settings, it is necessary to run Vision Language Models (VLMs) locally in the field while maintaining both high accuracy and computational efficiency.

However, applying foundation models designed primarily for cloud environments directly to the field presents challenges. As sensors increase in frequency and resolution, data transmission overhead grows, often degrading real-time performance. Furthermore, transmitting confidential or private information to external clouds raises security concerns, while unstable network environments hinder reliable operation. Consequently, deploying VLMs directly on local machines or edge devices is imperative. At the same time, because these edge environments typically suffer from limited computational resources and power constraints, it is essential to design highly efficient models without sacrificing accuracy.

If a VLM for autonomous devices outputs only natural-language answers without indicating the visual evidence for its decisions, tracing the root causes of malfunctions and ensuring operational safety becomes difficult. In real-world settings, model decisions also need to be communicated to human operators, recorded, and remotely shared in a usable format. In our target deployment environments in Japan, these outputs also need Japanese-language support so that they can be interpreted and relied upon in the field. For these reasons, grounded explanations and summaries are important requirements. Based on these requirements, PLaMo 2.1-VL emphasizes the following two fundamental capabilities:

\begin{itemize}
    \tightlist
    \item \textbf{Visual Question Answering (VQA):} the ability to process image and text inputs to generate a natural-language response.
    \item \textbf{Visual Grounding:} the ability to spatially identify the person or object referred to by a textual instruction within an image (including Referring Expression Comprehension (REC)).
\end{itemize}

Building on these capabilities, this report explores two real-world application scenarios: task estimation via tool recognition for factory task analysis, and infrastructure anomaly detection using patrol drones and surveillance cameras. In both scenarios, the objective is to adapt to specific field conditions while leveraging VQA and Visual Grounding as foundational tools.

To balance accuracy with the resource constraints of autonomous devices, we developed two variants of PLaMo 2.1-VL in a size range suitable for edge deployment: PLaMo 2.1-8B-VL, an 8B model for higher accuracy, and PLaMo 2.1-2B-VL, a 2B model designed for rapid prototyping and low-resource environments.

The main contributions of this report are as follows:
\begin{itemize}
    \tightlist
    \item We developed two lightweight VLMs, PLaMo 2.1-8B-VL and PLaMo 2.1-2B-VL, specifically designed for autonomous devices and optimized for practical Japanese-language operation.
    \item We enhanced VQA and Visual Grounding as core capabilities for autonomous systems, demonstrating their effectiveness across standard and translated Japanese benchmarks.
    \item To ensure practical field applicability and Japanese-language performance, we built a large-scale synthetic training data generation pipeline covering Visual Grounding, VQA, spatial understanding, counting, tool recognition, and difference detection.
    \item We evaluated two application tasks, factory task analysis and infrastructure anomaly detection, in both zero-shot and fine-tuned settings.
\end{itemize}

The remainder of this report is organized as follows: We first detail the model design and training methodology for PLaMo 2.1-VL. We then introduce the tasks and evaluation benchmarks, training data construction, and empirical results.

In this report, models with publicly available weights are referred to as ``open models'' for convenience. Furthermore, ``zero-shot'' denotes settings where no target-domain data is used for training or adaptation. Both PLaMo 2.1-8B-VL and PLaMo 2.1-2B-VL are publicly available for evaluation on Hugging Face.
\section{Model Design and Training Method of PLaMo 2.1-VL}
\label{sec:model-design}
\subsection{Overview of the Model Architecture}
\label{sec:model-architecture}
PLaMo 2.1-VL adopts a standard architecture similar to LLaVA~\cite{liu2023visual}. As with typical VLMs, PLaMo 2.1-VL consists of a large language model (LLM), an image encoder, and an image adapter that connects the two modalities. As the base LLM, we adopted PLaMo 2.1-8B or PLaMo 2.1-2B~\cite{networks2025plamo}, both of which were previously instruction-tuned via Direct Preference Optimization (DPO). This enables better instruction following. For the image encoder, we used SigLIP2~\cite{tschannen2025siglip} (\texttt{siglip2-so400m-patch14-384}~\cite{hf_siglip2_so400m_patch14_384}), and we utilized a straightforward multilayer perceptron (MLP) as the image adapter to map extracted visual features into an LLM-compatible representational space. Furthermore, to enhance the model's input representations, we adopted dynamic tiling, similar to NVIDIA Eagle 2~\cite{li2025eagle}, for image splitting and formatting. This allows images to be divided into appropriate patches, enabling the stable processing of varying resolutions and aspect ratios.
\subsubsection{Design Rationale and Trade-offs}
\label{sec:design-tradeoffs}
For selecting each architectural component, we prioritized not only benchmark performance but also hardware efficiency, particularly regarding GPU VRAM constraints during training. This section details our major design decisions and the underlying rationale.

\vspace{\baselineskip}
\noindent\textbf{Selection of the image encoder}: In a preliminary study, we evaluated several promising image encoders, including the SigLIP variant used in PaliGemma~\cite{beyer2024paligemma} and the ViT used in Qwen2.5-VL~\cite{qwen25_vl_2025}. Among these, SigLIP2 demonstrated the highest and most stable performance. Notably, beyond semantic understanding, it improved localization: the ability to accurately capture spatial layouts. Consequently, we hypothesize that SigLIP2 is useful for fine-grained tasks requiring local image understanding, such as semantic segmentation, object detection, and Visual Grounding/REC. One possible reason is that features corresponding to each patch represent localized information, which may make it easier to establish local correspondences (i.e., to identify specific referred regions) than with CLIP-style encoders, which tend to produce more global representations. We also observed that more stable region-target correspondence in Visual Grounding/REC was often associated with better spatial attention in VQA and may improve answer quality.

\vspace{\baselineskip}
\noindent\textbf{Selection of the image adapter}: Initially, we considered employing more complex adapters, such as Q-Former~\cite{li2023blip}, which dynamically select optimal image tokens. However, GPU VRAM was insufficient for a complex adapter without significantly reducing batch size. We therefore used a simpler MLP adapter, reducing memory overhead and securing a sufficient batch size to accelerate training. This efficiency allowed for more extensive hyperparameter tuning and iterative refinement, ultimately yielding higher overall benchmark scores.

\vspace{\baselineskip}
\noindent\textbf{Input representation}: For image splitting and formatting, we compared dynamic tiling against NaViT-style native resolution tokenization~\cite{navit_2023}, the method utilized in Qwen2.5-VL. Dynamic tiling was adopted as it consistently yielded superior results across our evaluation benchmarks.

\subsection{Training Method}
\label{sec:training-method}
Training a VLM requires endowing the LLM with visual comprehension while preserving its pre-existing conversational capabilities. Like LLaVA and Eagle 2, PLaMo 2.1-VL adopts a multi-stage training paradigm. Although we initially explored a three-stage configuration akin to Eagle 2, reproducing the performance gains of the final stage proved difficult. Consequently, we concentrated our development and optimization entirely on a two-stage design: Stage 1.0 and Stage 1.5.
\subsubsection{Stage 1.0: Alignment between Vision and Language}
\label{sec:stage10}
\textbf{Trainable components}: The image encoder and base LLM are entirely frozen; only the weights of the image adapter are updated.

\vspace{\baselineskip}
\noindent\textbf{Objective}: To align the vision and language modalities. The adapter is trained to map the visual features extracted by the image encoder into a representational space the LLM can comprehend. Because the base PLaMo 2.1 model is already instruction-tuned, freezing the LLM during this phase is crucial. This preserves the model's instruction-following behavior and safety guardrails while independently enhancing the image adapter's capabilities.

\vspace{\baselineskip}
\noindent\textbf{Training data}: We utilized web-scale image-caption datasets, mixing Japanese and English data at a ratio of $75:25$. Preliminary experiments indicated that incorporating 25\% English data improved downstream Japanese performance compared to strictly monolingual training. Given PLaMo 2.1's strong bilingual capabilities, Japanese and English tend to be closely aligned in its embedding space. Furthermore, English datasets typically feature higher annotation quality, serving as robust training signals that transfer effectively to Japanese tasks.

\subsubsection{Stage 1.5: Instruction Tuning and Visual Adaptation to Japanese}
\label{sec:stage15}
\textbf{Trainable components}: Low-Rank Adaptation (LoRA)~\cite{hu2022lora} is applied to the image encoder, image adapter, and base LLM, rendering all components trainable.

\vspace{\baselineskip}
\noindent\textbf{Objective}: To perform multimodal instruction tuning, enabling the model to handle complex image-related tasks and follow detailed instructions. Employing LoRA ensures training stability and memory efficiency while mitigating the catastrophic forgetting of pre-trained knowledge, thereby preserving the LLM's high-quality instruction-following capabilities.

\vspace{\baselineskip}
\noindent\textbf{Training data}: While maintaining the $75:25$ Japanese-to-English ratio, we significantly expanded task diversity. Alongside the caption data from Stage 1.0, we introduced curated datasets for VQA, Visual Grounding, open-vocabulary object detection, anomaly detection, and understanding of factory work.
\section{Tasks and Benchmarks}
\label{sec:tasks-benchmarks}
Building upon VQA and Visual Grounding as its core capabilities, PLaMo 2.1-VL tackles two real-world application tasks: factory task analysis and infrastructure anomaly detection. This section describes these four tasks and the benchmarks used to evaluate them. The first two evaluations measure the model's foundational capabilities, while the latter two are application tasks aimed at assessing practical field deployment.

Given PLaMo 2.1-VL's emphasis on practical Japanese-language operation, our evaluation prioritizes Japanese performance. For VQA, we use an established Japanese benchmark, and for Visual Grounding, we use a custom Japanese evaluation set in addition to an English benchmark. The methodology for preparing the Japanese evaluation set and our policy for translating the training data are detailed in Section~\ref{sec:data-translation}.
\subsection{VQA}
\label{sec:vqa-benchmark}
In VQA, we evaluate the model's ability to generate natural-language responses based on image content, given an image and a question as inputs. In this report, we utilize JA-VG-VQA-500~\cite{hf_ja_vg_vqa_500}, a Japanese benchmark. Compared with other Japanese VQA datasets, this benchmark contains a larger number of samples and is better suited for quantitatively evaluating fundamental visual recognition capabilities in Japanese, including object recognition, attribute recognition, counting, and spatial reasoning.

We employ multiple evaluation metrics: ROUGE-L~\cite{lin2004rouge}, LLM-as-a-judge~\cite{zheng2023judging}, and English Likert LLM judge~\cite{sasagawa2025constructing}. Because ROUGE-L is fundamentally a string-overlap metric, it is insufficient for evaluating responses that are semantically correct but phrased differently, or for capturing nuances in answer granularity (e.g., ``dog'' versus ``brown dog''). Therefore, we complement this with LLM-based metrics to assess semantic validity. Specifically, the LLM-as-a-judge provides a correctness-oriented assessment~\cite{zheng2023judging}, whereas the English Likert LLM judge provides a finer-grained evaluation on a five-point scale~\cite{sasagawa2025constructing}.
\subsection{Visual Grounding}
\label{sec:vg-benchmark}
Visual Grounding evaluates the ability to localize the specific region within an image referred to by a text prompt. In this report, we focus on Referring Expression Comprehension (REC) as a representative setting, utilizing the English Ref-L4 benchmark~\cite{hf_ref_l4} and the Japanese Ref-L4 benchmark prepared by Preferred Networks, Inc. (PFN)~\cite{github_ja_ref_l4}.

Historically, representative REC benchmarks have included RefCOCO, RefCOCO+, and RefCOCOg~\cite{yu2016modeling}. However, because these benchmarks consist of relatively short phrases, performance scores have begun to saturate for recent high-performance VLMs. In contrast, Ref-L4 assigns multiple captions to a single ground-truth bounding box, incorporating detailed explanations and diverse paraphrases. These natural language expressions are long, complex, and highly varied, ensuring the benchmark remains sufficiently challenging even for state-of-the-art VLMs.

To ensure fair evaluation in Japanese, it is necessary to produce translations that preserve the inherent difficulty and instructional structure of the English version. Therefore, PFN constructed a Japanese version of Ref-L4 that rigorously maintains the structural complexity of the original captions. Further details are provided in Section~\ref{sec:ref-l4-japanese}.
\subsection{Factory Task Analysis}
\label{sec:factory-task-analysis}
As one of the primary intended use cases for PLaMo 2.1-VL, this report addresses factory task analysis. In industrial environments, accurately inferring a worker's current task solely from their posture and surroundings is challenging. However, many tasks are strongly associated with specific tools. For instance, recognizing with high accuracy whether a worker is using a torque wrench or a screwdriver allows the model to identify the task with significantly higher confidence. Therefore, we adopt a setting in which work tasks are estimated using tool recognition as a primary cue, evaluating how well the model can identify task content from field images. Here, we focus specifically on identifying the tool categories that directly contribute to task analysis.

For evaluation, we use an in-house factory task analysis benchmark. This benchmark consists of images of workers performing ten types of tasks in an actual factory, requiring the model to classify the depicted task into one of ten classes. During inference, we utilize a prompt that provides the visual appearance information of tools as hints, leveraging the strong association between tasks and specific tools. Because providing only tool names can make recognition unstable for highly specialized tools, this report also incorporates visual cues, such as shape, color, and characteristic components, to assist with visual identification. The model is strictly instructed to output only the corresponding task label. A specific example of this prompt is provided in Appendix Section~\ref{sec:appendix-factory-prompt}.
\subsection{Infrastructure Anomaly Detection}
\label{sec:infra-anomaly-detection}
Anomaly detection for infrastructure facilities is a highly variable domain: appearances and installation environments differ drastically between facilities, and public datasets or web-sourced images are scarce, making large-scale training data collection difficult. Consequently, approaches must be capable of operating in zero-shot settings or with minimal data, without being tailored to a specific facility. In this development, we leverage the operational reality that patrol drones and surveillance cameras repeatedly capture the same locations. We adopt a comparative methodology that detects anomalies by contrasting a reference image (normal state) with a target image (current state). Specifically, we treat the reference image as the baseline and extract changes indicative of anomalies based on the visual differences. The anomalies we consider include damaged or missing equipment, the appearance of objects that should not be present, liquid leakage traces, and changes in the open/closed states of doors or valves.

However, in drone-based imaging, even when images are captured along the same preplanned flight route, the imaging position and camera angle never match perfectly due to GPS/flight control errors and disturbances such as wind. Even static surveillance cameras experience micro-vibrations and viewpoint shifts. Furthermore, variations in lighting and weather mean that simple pixel-difference comparisons are easily dominated by apparent differences caused by misalignment and illumination changes, rendering detection highly unstable.

Therefore, our approach trains the VLM to compare two images and extract semantically meaningful changes indicative of anomalies, rather than these apparent differences. In addition, given the impracticality of collecting large quantities of real-world images of infrastructure anomalies, we train the model using general synthetic data to teach the fundamental mechanics of two-image difference detection. This strategy ensures that the model can be deployed to new sites without excessive dependence on a particular environment. The model outputs bounding boxes indicating the positions of anomaly candidates, along with anomaly labels assigned to each box. The synthetic data used for training difference detection is described in Section~\ref{sec:anomaly-synthetic-data}.
\subsubsection{Construction of the Evaluation Dataset and Pair Images}
\label{sec:eval-dataset-pair-images}
To evaluate anomaly detection performance, we utilize drone data captured at multiple facilities within a power plant. To verify the model's ability to handle realistic anomalies, physical anomalies were artificially staged at these facilities and photographed using a camera-equipped drone. Imaging was conducted across the facilities over three days, with the drone flown repeatedly along a predefined flight route for each facility. Images were captured under two conditions: reference images without anomalous objects (normal state), and target images with artificially placed anomalous objects (abnormal state). To ensure diversity in the evaluation data, the anomalous objects were rearranged and varied in type during each capture session. Capturing data across three days also provided natural variation in lighting and weather.

Annotations were applied to the target images, assigning a bounding box and one of fourteen anomaly labels to each anomalous object. The 14 categories include: plastic bottles, empty cans, tools, garbage bags, traffic cones, gloves, helmets, bird nests, umbrellas, towels, protective tape, people, open doors (e.g., on electrical panels), and water leakage. The label definitions were designed with actual operational semantics in mind; for example, for open doors, the bounding box is assigned strictly to the open door itself, and for water leakage, the bounding box encapsulates the wet regions rather than the source of the leak. Approximately 40\% of the target images contain at least one anomaly. While the number of anomalous objects per image is typically one, some images contain between two and six. Additionally, because anomalous objects can appear as small as few tens of pixels, this setup also rigorously evaluates small-object detection capabilities.

\begin{figure}[tb]
    \centering
    \includegraphics[width=0.99\linewidth]{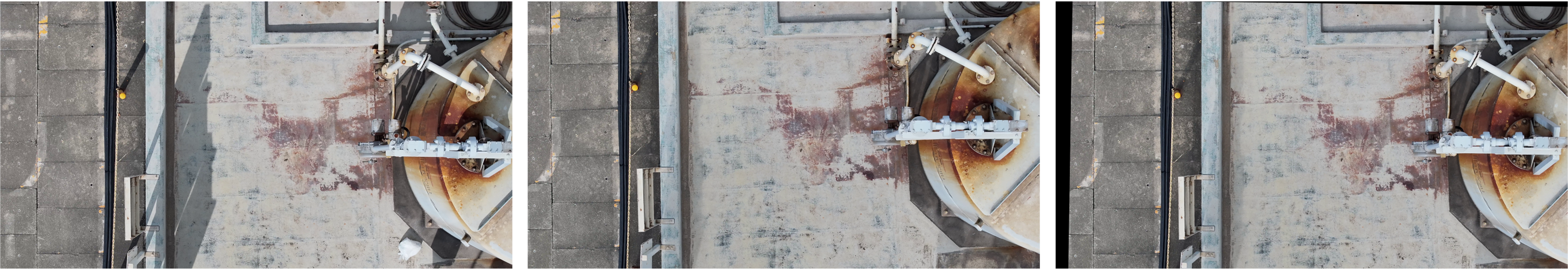}
    \caption{Example of pair image generation. (a) Target image, (b) selected raw reference image, and (c) reference image post-homography transformation.}
    \label{fig:pair_image_creation}
\end{figure}

The input for difference detection is a single composite image created by placing the target image (abnormal state) on the left and the reference image (normal state) on the right. As noted above, the imaging positions and camera angles never align perfectly. Although it is possible to input misaligned images directly into the VLM, forcing the model to implicitly learn spatial alignment increases the training burden and makes it more susceptible to apparent differences. Therefore, as a preprocessing step, for each target image, we first retrieve candidate reference images captured from nearby positions and then select the visually most similar one and mathematically align it before forming the composite pair.

The procedure is as follows: First, using telemetry data (latitude, longitude, altitude) recorded during capture, we select multiple candidate reference frames based on the drone's spatial proximity. Multiple candidates are necessary because the closest spatial position does not always guarantee the closest camera angle, due to variations in drone posture. Next, we perform image matching using SIFT~\cite{lowe2004distinctive} and LightGlue~\cite{lindenberger2023lightglue} on each candidate frame. The image with the smallest average image-plane distance between corresponding feature points is selected as the optimal reference image. Finally, this reference image is corrected using a homography transformation estimated from the matching results, reducing viewing angle discrepancies and positional misalignment. The final input image is then created by concatenating the target and transformed reference images. Figure~\ref{fig:pair_image_creation} illustrates (a) a target image, (b) the selected raw reference image, and (c) the reference image post-homography transformation. As shown in (b), the raw reference image exhibits positional shift and slight counterclockwise rotation relative to the target. In contrast, in (c), the feature points in the reference image have been mathematically warped to align with the target image, effectively correcting the positional and angular misalignment. Simultaneously, minor scale discrepancies are also corrected. Note that because reference images are retrieved based on the target image, pairs with significant appearance differences may occasionally be generated. Examples include heavy blurring, drastic appearance changes, or severe angle mismatches caused by drone posture shifts despite close spatial proximity. We treat these challenging pairs as an inherent part of the evaluation conditions, accurately reflecting the variations expected in real-world drone operations.
\subsubsection{Inference Method and Evaluation Metrics}
\label{sec:anomaly-inference-eval}

\begin{figure}[tb]
    \centering
    \includegraphics[width=0.90\linewidth]{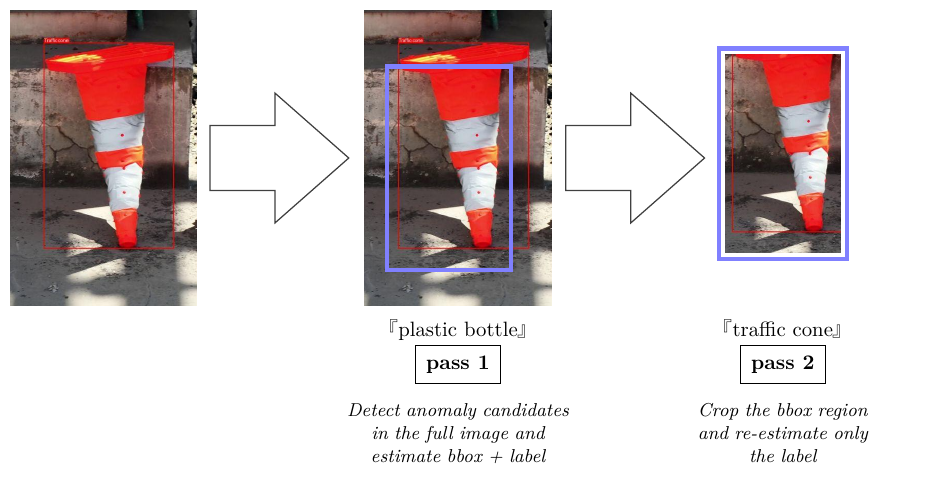}
    \caption{Two-pass inference flow for anomaly detection. In pass 1, anomaly candidates are searched over the whole image and bounding boxes and labels are estimated. In pass 2, the bounding box regions obtained in pass 1 are cropped and only the labels are re-estimated. (The red boxes in the figure indicate ground-truth bounding boxes and labels.)}
\label{fig:2stage_inference}
\end{figure}

In anomaly detection, it is critical not only to flag the presence of an anomaly, but also to output explainable visual evidence of \textit{where} and \textit{what} has changed. To achieve this, our method adopts the two-pass inference procedure illustrated in Figure~\ref{fig:2stage_inference}. In the first stage (pass 1), the entire image is used as input to search for anomaly candidates, jointly estimating bounding boxes and preliminary labels. In the second stage (pass 2), the bounding box regions obtained in pass 1 are cropped, and these localized region images are used as inputs to re-estimate the labels. When bounding boxes and labels are estimated jointly from the full image, contextual information (such as background clutter) often acts as noise, destabilizing label classification. By localizing the anomaly candidates first and re-estimating the labels using tight crops, we aim to significantly improve both accuracy and stability.

To evaluate anomaly detection performance under realistic conditions, we constructed a benchmark using the power plant data described previously. The evaluation dataset consists of 1,200 samples in total, including 400 samples containing one or more anomalies from each of three distinct plant locations. Each sample is an image pair (target and reference) with a concatenated resolution of $7680 \times 2160$ pixels, classified across 14 anomaly labels.

With real-world deployment in mind, we evaluate the model under two operational scenarios. The first scenario isolates the model's ability to identify the spatial positions of anomaly candidates (bbox only). This reflects an operational flow where an operator receives an alert, visually inspects the flagged coordinates, and manually determines the response. The second scenario evaluates the model's ability to identify both the position and the exact nature of the anomaly (bbox + label), enabling more efficient responses tailored to the anomaly type, as well as cross-facility aggregation and reporting.

As our primary evaluation metric, we calculate the F1-score for each image pair individually and report the macro-average across all pairs. Computing a global bounding-box-level metric would disproportionately weight images containing numerous anomalous objects. By computing an F1-score per sample, we evaluate image-level detection performance while treating all samples equally. The strict mathematical definition of this evaluation metric and the matching criteria are detailed in Appendix Section~\ref{sec:appendix-anomaly-metrics}.
\section{Data Synthesis}
\label{sec:data-synthesis}
One important factor determining the performance of PLaMo 2.1-VL is the quality and composition of its training data. Publicly available datasets alone are insufficient to satisfy real-world operational requirements, which demand handling domain-specific images, precise local visual understanding, and fine-grained instruction adherence. Consequently, we positioned synthetic data generation as a central pillar of this development, constructing large-scale training data in-house.

The foundation of this training data is a collection of Japanese web images accumulated through web crawling conducted at PFN. First, to explicitly define the scope of our synthetic data, we automatically classified the crawled images into ``natural images'' (photographs) and ``document images'' (scans, printed pages, text documents, etc.). Because the primary tasks targeted in this development (such as Visual Grounding, VQA, tool recognition, and difference detection) rely heavily on object recognition, spatial relations, and local visual features, our synthesis pipeline focused exclusively on natural images. Document images fall outside the scope of this report.

Starting from these natural images, we automatically generated task-specific supervision signals to construct our datasets. At the onset of the project, synthesis was conducted using Qwen2.5-VL-32B-Instruct~\cite{hf_qwen25_vl_32b_instruct}. Following the release of Qwen3-VL-235B-A22B-Instruct~\cite{bai2025qwen3}, we transitioned to using it as our primary generation model. This section first describes the synthesis of training data for the foundational capabilities of Visual Grounding and VQA, followed by the generation of data for factory task analysis and anomaly detection.
\subsection{Synthetic Data for Visual Grounding}
\label{sec:vg-synthetic-data}
Visual Grounding is the task of localizing a region within an image based on a provided textual referring expression, typically outputting a bounding box or segmentation mask. When deploying VLMs in the field, it is crucial not only that the model provides the correct answer, but also that it can visually explain the evidence behind its decision. Strengthening Visual Grounding is therefore expected to enhance both the reliability of VQA and the safety and interpretability of downstream application tasks.

Because PLaMo 2.1-VL is designed to output bounding boxes for Visual Grounding, we synthesized data specifically tailored for bounding-box supervision. This synthesis pipeline consists of two main stages:

\vspace{\baselineskip}
\noindent\textbf{Generation of referring expressions and labels:} First, Qwen3-VL-235B-A22B-Instruct is prompted to generate referring expressions targeting specific instance sets within the image (e.g., ``black cars''). Categories are not predefined; instead, the natural language strings output by Qwen3-VL-235B-A22B-Instruct serve simultaneously as the referring expressions and the class labels. This open-vocabulary approach allows us to broadly cover both general objects and domain-specific concepts.

\vspace{\baselineskip}
\noindent\textbf{Region extraction and conversion to bounding boxes:} Next, the generated referring expressions are input into SAM3~\cite{carion2025sam3segmentconcepts} to extract the target regions. To utilize these as training supervision signals, the segmentation masks output by SAM3 are converted into bounding boxes. When a referring expression targets multiple instances, all corresponding bounding boxes are retained. This multi-box output is also leveraged later for training counting tasks. However, depending on the complexity of the referring expression, SAM3 occasionally fails to accurately recognize the target, returning an empty region. As an exception-handling fallback in these cases, Qwen3-VL-235B-A22B-Instruct is prompted to output the bounding boxes directly, acting as a substitute for SAM3.
\subsubsection{Transitioning from Single-Bounding-Box Bias to Multi-Instance Support}
\label{sec:single-bbox-bias}
In our initial attempts to improve Visual Grounding, training relied on datasets where the generated referring expression uniquely identified a single specific instance, paired with exactly one ground-truth bounding box. However, models trained on this data absorbed the distribution bias, developing a tendency to always output exactly one bounding box. Consequently, when presented with referring expressions common in real-world operations that inherently target multiple instances (e.g., ``all the apples on the table''), the model would often incorrectly return only a single bounding box, severely limiting its generality.

To resolve this, we revised our data synthesis policy to intentionally generate referring expressions targeting multiple instances, explicitly injecting them into the training data. This shifted the model's output distribution, enabling it to naturally handle both single instances and instance sets. Furthermore, as we refined our region extraction process, empirical observations revealed that bounding boxes derived from SAM3 masks were consistently more accurate and stable than those directly output by Qwen3-VL-235B-A22B-Instruct. We therefore adopted a hybrid architecture: the primary workflow relies on SAM3 for localization, while direct bounding box prediction via Qwen3-VL-235B-A22B-Instruct is strictly reserved as an exception-handling mechanism for when SAM3 fails.
\subsection{Synthetic Data for VQA}
\label{sec:vqa-synthetic-data}
For synthetic VQA data, we automatically generated question-answer pairs grounded in visual content. A critical priority here was to avoid introducing questions that could be answered from prior common knowledge alone, or ambiguous questions subject to unstable interpretation. Therefore, we designed a generation process ensuring that the resulting QA pairs cannot be accurately answered without direct reference to the image.

Specifically, rather than generating QA directly from the image in a single pass, we employed a two-stage configuration: caption generation followed by text-based QA generation. For the first stage, we prompted Qwen2.5-VL-32B to describe only verifiable facts directly observable in the image, forbidding speculative language or subjective interpretation. In the second stage, we used the text-only LLM Qwen3-32B (not a VL model)~\cite{hf_qwen3_32b}, instructing it to generate questions and answers based only on the content provided in the caption.
\subsubsection{Spatial Understanding}
\label{sec:spatial-understanding}
For Spatial Understanding, we synthesized VQA data that queries the relative positions of two instances in an image (e.g., left-right, above-below, front-back). For this task, labels become unreliable when speculation or interpretation is mixed with factual content. Therefore, it is critical that the supervision signals capture only relations that can be stated from visual evidence alone.

To achieve this, we decoupled relation extraction from QA generation, synthesizing the data across three stages:

\vspace{\baselineskip}
\noindent\textbf{Enumeration of instances in the image:} First, Qwen3-VL-235B-A22B-Instruct generates a structured list (e.g., in JSON) of objects visible in the image. The prompt enforces rules: ``do not speculate'' and ``write only visible facts.'' This enumeration is then used as the candidate pool for relations extraction.

\vspace{\baselineskip}
\noindent\textbf{Structured extraction of relations between instances:} The enumerated instance list is fed back into the model, which is prompted to extract valid relationships and output them as subject/relation/object triplets. The relations are limited to those that can be verified from visual evidence. These include spatial relations (left of/right of/above/below/in front of/behind) as well as other visually clear relations (facing/holding/connected to/carrying/sitting on). This triplet format makes the spatial relationship explicit. For example, given the subject ``the person on the right side of the image,'' the relation ``left of,'' and the object ``the chair near the center,'' the triplet asserts: ``the person on the right side of the image is to the left of the chair near the center.'' The subject and object phrases may contain contextual cues to help identify the targets, but the relation itself describes only the relation between them. This makes the spatial evidence explicit for the later QA generation stage.

\vspace{\baselineskip}
\noindent\textbf{QA generation from extracted relations:} Finally, the structured triplets are used to generate multiple VQA-style questions and answers. A single triplet can spawn diverse paraphrases and question formats, such as ``Is A to the left of B?'' or ``What is to the right of B?''

We found that when relation extraction and QA generation were performed simultaneously, the model tended to overlook instances, generate mismatched QA pairs, or introduce speculative relations. By isolating the process into three simplified stages, we reduced missed relations and QA inconsistencies.
\subsubsection{Counting}
\label{sec:counting-synthetic-data}
In visual counting tasks, even a minor error in the count label becomes training noise. To improve label reliability, we kept only samples for which Qwen3-VL-235B-A22B-Instruct and SAM3 produced the same count.

The synthesis process is as follows: First, Qwen3-VL-235B-A22B-Instruct generates referring expressions targeting a set of specific instances. To avoid generating simple prompts (e.g., ``car''), the model is instructed to generate visually descriptive phrases (e.g., ``black cars,'' ``parked cars''). This yields diverse counting signals from the same underlying image and category.

Next, this referring expression is evaluated by both Qwen3-VL-235B-A22B-Instruct and SAM3. Qwen3-VL-235B-A22B-Instruct directly outputs a numerical count (\texttt{count\_qwen}). Simultaneously, SAM3 extracts the target regions, converts the masks to bounding boxes, and the number of resulting boxes is tallied (\texttt{count\_sam}). Only samples where \texttt{count\_qwen} equals \texttt{count\_sam} are retained in the dataset, filtering out erroneous labels. We call this mechanism the \textbf{dual-estimation agreement gate}.

Additionally, because the model generates its own referring expressions, it occasionally targets uncountable concepts (e.g., ``sky''). We therefore apply a \textbf{countability filter} and keep only referring expressions that refer to discrete, countable entities.

During initial prototyping, generating Counting QA directly from images occasionally resulted in data leakage, producing flawed questions such as ``How many three cars are there?'' We first attempted a staged generation approach (enumerating instances, estimating the count, then rewriting into VQA format) to mitigate this. While this reduced prompt leakage, enumeration-based noise persisted. We therefore replaced this approach with the dual-estimation agreement gate and the countability filter.
\subsection{Synthetic Data for Factory Task Analysis}
\label{sec:factory-synthetic-data}
In analyzing the workflow of factory personnel, tool recognition is critical. To build this capability, we automatically synthesized Visual Grounding training data by extracting tool-containing images from our internally collected Japanese web-crawled images. First, Qwen3-VL-235B-A22B-Instruct, which demonstrates high proficiency in identifying both common and specialized industrial tools, was used to detect tools and output bounding boxes. Next, these bounding boxes were fed back into Qwen3-VL-235B-A22B-Instruct to generate corresponding referring expressions (captions), thereby synthesizing REC data.

However, without further filtering, this process occasionally generates referring expressions that fail to uniquely identify the target tool in the image. To mitigate this, we introduced a \textbf{self-consistency filter}. The generated caption and the source image are fed back into the model to re-predict a bounding box, and the Intersection over Union (IoU) between this new prediction and the original bounding box is calculated. By discarding samples with a low IoU, we ensure that the referring expression reliably and uniquely specifies the target object before incorporating it into the final dataset.

\subsection{Synthetic Data for Anomaly Detection}
\label{sec:anomaly-synthetic-data}
A challenge in infrastructure anomaly detection is that public data is scarce, while real-world equipment and environments are diverse. This makes large-scale training on real images difficult. To address this, we created synthetic data using instance masks from the Open Images Dataset V7~\cite{openimages_v7}. We used this data for two purposes: to increase the amount of training data and to teach the model the mechanics of difference detection by comparing a reference image with a target image.
\subsubsection{Basic Design of the Synthetic Data}
\label{sec:anomaly-synth-basic}
\begin{figure}[tb]
\centering
    \includegraphics[width=0.99\linewidth]{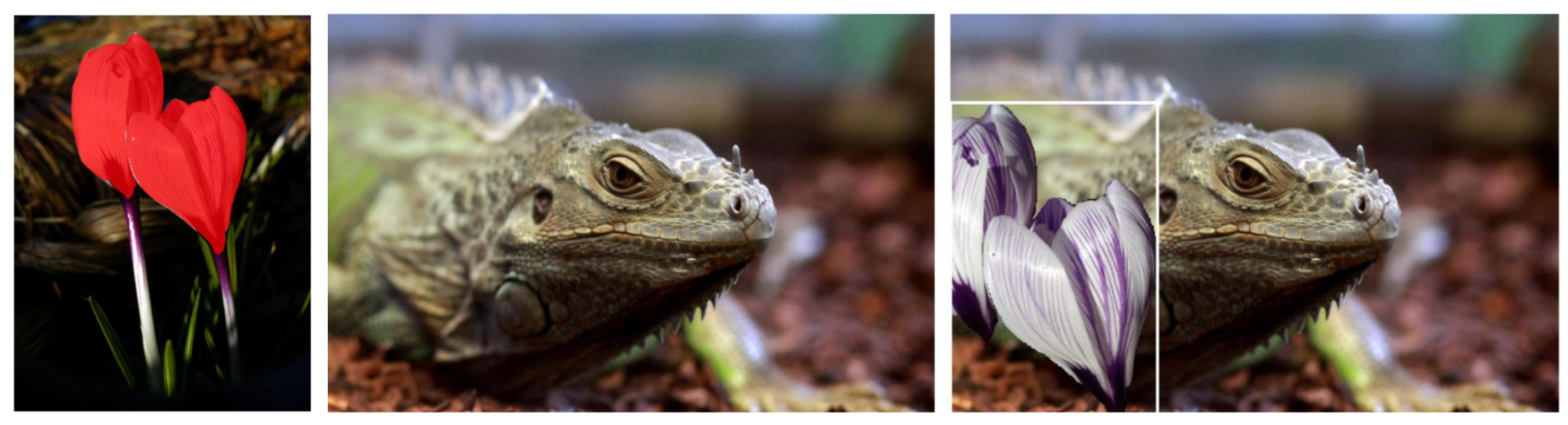}
    \caption{Example of a synthesized anomalous image created for difference detection. From left to right, the figure shows the source object image, the destination image for overlay, and the anomalous image generated by overlaying the object.
    (Left) Source image: Tom May, ``P1210645-a'', Flickr, CC BY 2.0 Work page:~\cite{flickr_p1210645a}. License:~\cite{cc_by_20}. Modification: The image has been processed by overlaying the instance mask in red.
    (Center) Source image: Mohd Fazlin Mohd Effendy Ooi, ``Sunway Lagoon'', Flickr, CC BY 2.0 Work. page:~\cite{flickr_sunway_lagoon}. License:~\cite{cc_by_20}.
    (Right) Image generated by overlaying the image cut from the left image onto the center image. 
    }
    \label{fig:synthesized_anomalous_image}
\end{figure}

To synthesize an anomalous image, an object region is cropped from a source image using an instance mask and overlaid onto a destination image. This artificially simulates the appearance of a foreign object that would not normally be present (Figure~\ref{fig:synthesized_anomalous_image}). The pre-overlay destination image is used as the reference image, the post-overlay image is used as the target image, and the bounding box of the injected object is used as the supervision signal. This procedure generates the image pairs for difference detection and the bounding boxes for localization.

\begin{figure}[tb]
    \centering
    \includegraphics[width=0.99\linewidth]{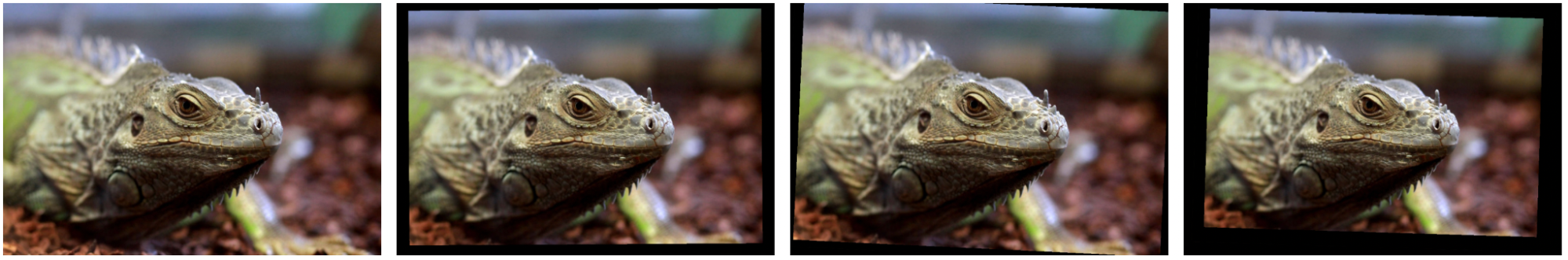}
    \caption{Example of adding geometric perturbations. This figure shows an example in which mild geometric transformations were added to the reference image to simulate differences in imaging position and camera angle.
    (Leftmost) Source image: Mohd Fazlin Mohd Effendy Ooi, ``Sunway Lagoon'', Flickr, CC BY 2.0 Work page:~\cite{flickr_sunway_lagoon}. License:~\cite{cc_by_20}.
    (Center-left, center-right, and rightmost) Affine transformations of the
    leftmost image.
    }
    \label{fig:geometric_transformations}
\end{figure}

In real-world operations, reference and target images are captured at different times, meaning that imaging positions and camera angles do not match perfectly, resulting in parallax and misalignment. Therefore, during synthesis, we apply mild geometric perturbations to the reference image to simulate variations in camera angle and positional shift (Figure~\ref{fig:geometric_transformations}). This forces the model to learn to extract semantic, anomaly-related differences while ignoring pixel-level differences caused by misalignment.

\subsubsection{Prompt Design for Difference Detection}
\label{sec:diff-detection-prompt}
During training, prompts were automatically generated from multiple templates, simultaneously teaching the model to follow instructional formatting, compare the two images, and detect anomalies. These prompts were designed so that both the comparison task and the anomaly classes to detect could be specified explicitly in text. For example, a prompt might instruct the model to locate objects present in the left image (target) but absent in the right image (reference), given a specific list of candidate classes:

\begin{itemize}
    \tightlist
    \item ``Find the bounding boxes of objects that are present in the left image but missing in the right image: \texttt{[apple, book, camera]}''
    \item ``Detect all items of the specified classes that appear in the left frame but are not found in the right frame: \texttt{[handbag, tie, license plate]}''
\end{itemize}

Including the target classes in the prompt makes the detection target explicit. Crucially, the candidate lists provided in the prompts were designed to include both the actual injected anomalies and several ``dummy'' classes that were absent from the image pair. This strategy prevents the model from unconditionally over-detecting items simply because they appear in the prompt text, teaching it to output only what is visually verifiable. Consequently, the model learns to use the prompt to guide detection without hallucinating dummy classes.

Finally, to investigate the effect of instruction granularity during training, we trained separate model variants using different instruction styles:

\begin{itemize}
    \tightlist
    \item \textbf{Specific instruction:} Enumerates the detection targets using exact class names (e.g., ``apple, book'').
    \item \textbf{Abstract instruction:} Generalizes the detection targets using broader terms (e.g., ``object'') rather than specific class names.
\end{itemize}

Evaluation showed that while models trained under both training schemes could detect anomalies when given abstract instructions during inference, the model trained with abstract instructions had a higher false-positive rate. For instance, it tended to detect objects outside the 14 target classes, such as detecting a tablet held by a worker who was the actual intended anomaly target. This made head-to-head comparisons against specific-instruction inference unfair. In addition, when both models were evaluated using specific instructions, the model trained with specific instructions demonstrated superior overall performance. Consequently, we adopted the specific-instruction training paradigm for the final model.
\section{Data Translation and Preparation of Japanese Evaluation Sets}
\label{sec:data-translation}
PLaMo 2.1-VL is designed for practical Japanese-language operation. This requires translating English-centric public and synthetic datasets into Japanese. It also requires evaluation sets capable of measuring Japanese performance. This section first describes the policy and implementation details for translating the training data. It then explains how the Japanese version of the Ref-L4 benchmark, used to evaluate Visual Grounding, was constructed.
\subsection{Translation of Training Data}
\label{sec:training-data-translation}
The majority of publicly available image-text datasets are in English. To ensure strong Japanese-language performance, we translated the English text (e.g., captions and instructions) within these datasets. Furthermore, because synthetic data generation via existing LLMs and VLMs is generally more stable and yields higher quality in English, our standard pipeline involved generating synthetic data in English prior to Japanese translation.

For translation, we utilized the PLaMo translation model developed by PFN~\cite{hf_plamo2_translate}. This model is available in two main variants: a base model and a post-trained model. Both allow the translation policy to be adjusted via few-shot examples, but they exhibit distinct characteristics. The post-trained model generally produced better translation quality, but it followed the provided examples less closely. Conversely, the base model exhibited slightly lower standalone translation quality but followed the structure of few-shot examples much more rigidly.

When translating training data, simply generating natural Japanese is insufficient. Some datasets intentionally contain unnatural phrasing or a wide range of expression styles. If the translation model rewrites them into smoother Japanese, the original phrasing and variation are lost. Furthermore, directly translating literal strings that appear inside the image (such as signs or labels) causes a mismatch between the image and the text.

To address these issues, we designed our few-shot examples around two principles: (1) maintain the wording and style of the source text as much as possible without over-naturalizing it, and (2) leave literal strings enclosed in double quotation marks untranslated. Based on these principles, we defaulted to the base model because it followed the few-shot examples closely. However, the base model occasionally suffered from omissions in long inputs or hallucinated additions. To mitigate this, we leveraged the heuristic that input and output token counts remain similar during translation. When the output token count fell outside a predetermined range relative to the input, we treated the result as potential mistranslation and reprocessed it using the post-trained model. This hybrid approach combined the base model's structural preservation with the post-trained model's translation quality, stabilizing and improving the overall translation pipeline.

As an additional refinement, we incorporated overall image descriptions (generated by Qwen2.5-VL-32B-Instruct or Qwen3-VL-235B-A22B-Instruct) along with other texts attached to the same image, as contextual input during translation. This supplementary context helps disambiguate homographs (e.g., distinguishing a baseball ``bat'' from the flying mammal).
\subsection{Construction of the Japanese Version of Ref-L4}
\label{sec:ref-l4-japanese}
\begin{figure}[tb]
    \centering
    \includegraphics[width=0.50\linewidth]{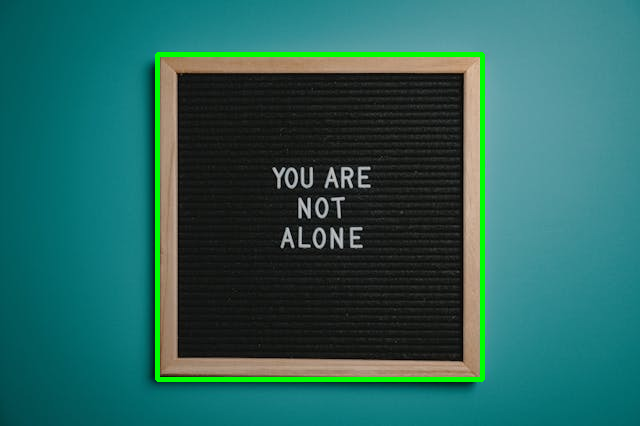}
    \caption{Example image with enclosed text for translation.
    Source: Pexels License:~\cite{pexels_license}. URL:~\cite{pexels_2821220}.
    For inclusion in this report, an additional bounding-box overlay has been
    added.
    }
    \label{fig:example_translation}
\end{figure}

To construct the Japanese version of Ref-L4, we utilized the PLaMo translation model described above. As with the training data, the primary objective was not only to produce natural Japanese, but also to preserve both the referential difficulty and the structural complexity of the original English Ref-L4 benchmark. Ref-L4 features many captions containing literal strings present in the image, as well as long referring expressions with multiple modifying clauses. Standard machine translation tends to break both visual consistency and original sentence structure in these cases. Below, we outline representative translation challenges inherent to Ref-L4 and our corresponding solutions. Note that the images and captions presented below are illustrative examples modeled after Ref-L4 examples, rather than exact excerpts.

\vspace{\baselineskip}
\noindent\textbf{Captions containing strings appearing in the image:} In tasks requiring the detection of text regions within an image, captions frequently enclose the literal text present in the image within double quotation marks. Consider the following caption:

\begin{itemize}
    \tightlist
    \item Source text: \texttt{A\ framed\ sign\ reading\ "YOU\ ARE\ NOT\ ALONE".}
\end{itemize}

Here, \texttt{"YOU\ ARE\ NOT\ ALONE"} is a string written in English within the image (see Figure~\ref{fig:example_translation}). Translating this string into Japanese would create a mismatch between the text prompt and the visual evidence.

Because general machine translation often erroneously translates these literal strings, we provided the PLaMo translation model with specific few-shot examples to induce it to leave quoted strings untouched:

\begin{itemize}
    \tightlist
    \item Source example: \texttt{T-shirt\ with\ "New\ York"\ printed\ on\ it}
    \item Translation example: \texttt{「New\ York」とプリントされたTシャツ}
\end{itemize}

By supplying these examples, the model preserves the quoted strings in actual captions. Consequently, the original sample above is translated into a Japanese expression equivalent to \texttt{"YOU\ ARE\ NOT\ ALONE"と書かれた額縁入りの看板}.

\vspace{\baselineskip}
\noindent\textbf{Phrases containing multiple modifying clauses:} Another challenge involves long referring expressions in which multiple clauses modify a single noun phrase. Consider the following:

\begin{itemize}
    \tightlist
    \item Source text: \texttt{A\ brown\ and\ white\ spotted\ long-haired\ dog\ watching\ another\ dog\ and\ a\ human,\ positioned\ closest\ to\ the\ foreground.}
\end{itemize}

In this sentence, two modifying clauses (\texttt{watching\ another\ dog\ and\ a\ human} and \texttt{positioned\ closest\ to\ the\ foreground}) both modify the noun phrase \texttt{A\ brown\ and\ white\ spotted\ long-haired\ dog}. Standard machine translation tends to paraphrase them into natural Japanese predicate sentences, which breaks the chained modifying structure. However, in Ref-L4, this structure is an important cue for identifying the target. Therefore, we provided few-shot examples with a similar structure to induce translations that preserve the noun-phrase-centric structure:

\begin{itemize}
    \tightlist
    \item Source example: \texttt{The\ man\ in\ blue\ with\ both\ hands\ raised,\ standing\ at\ the\ farthest\ end\ of\ the\ screen.}
    \item Translation example: \texttt{画面の最も遠い端に立っている、両手を上げた青い服を着た男性。}
\end{itemize}

This prompting strategy ensures the generation of Japanese translations that retain the complex modifying clauses. Thus, the dog example can be translated into a Japanese expression equivalent to \texttt{the\ brown-and-white\ spotted\ long-haired\ dog,\ closest\ to\ the\ foreground,\ watching\ another\ dog\ and\ a\ human}.

\vspace{\baselineskip}
\noindent\textbf{Use of context and selection among translation candidates:} To further stabilize translation quality, the output can be partially controlled by providing contextual examples or explanatory text. During translation, we provided overall image descriptions (generated by Qwen3-VL) and other bounding-box captions from the same image as context. This provides the necessary information for accurate word-sense selection and target identification. However, because providing context does not universally improve translation quality (and can sometimes even degrade it), we generated multiple translation candidates by varying the combinations of examples and model settings. We then computed the semantic similarity to the source text using the embedding model \texttt{plamo-embedding-1b}~\cite{hf_plamo_embedding}, selecting the candidate with the highest similarity score.

Ultimately, when constructing the Japanese version of Ref-L4, we prioritized visual consistency, preservation of structure, and the maintenance of referential difficulty. However, because both the translation and candidate selection rely on LLM-based pipelines, not every sample is guaranteed to have been translated perfectly. Detailed evaluation results on this benchmark are provided in Section~\ref{sec:results-vg}.
\section{Results}\label{sec:results}
This section presents the evaluation results for PLaMo 2.1-VL. We first report performance on Japanese and English benchmarks for the core capabilities of VQA and Visual Grounding. We then present the results for our two real-world application tasks, factory task analysis and infrastructure anomaly detection, discussing both the potential and the remaining challenges of zero-shot deployment. For our baselines, we primarily adopt strong open models of similar size available at the time of evaluation, including larger models where necessary for context. Details regarding the benchmarks and evaluation metrics for each task are provided in Section~\ref{sec:tasks-benchmarks}.
\subsection{VQA}
\label{sec:results-vqa}
To evaluate VQA capabilities, we utilized the JA-VG-VQA-500 benchmark and compared models across multiple metrics: ROUGE-L, LLM-as-a-judge, and English Likert LLM judge. Our baselines included (1) Asagi-14B~\cite{hf_asagi_14b}, which held the highest score on JA-VG-VQA-500 among open models at the start of this project (April 2025), and (2) Qwen3-VL-8B-Instruct~\cite{hf_qwen3_vl_8b_instruct}, a high-performing open model in a similar size class to PLaMo 2.1-8B-VL. Note that Qwen3-VL-235B-A22B-Instruct was excluded from this comparison because it tended to generate overly specific answers that misaligned with the benchmark's intended granularity, resulting in unfairly low scores.

Table~\ref{tab:VQA_evaluation_results} shows the results. PLaMo 2.1-8B-VL achieved higher scores than the baselines across multiple metrics. Furthermore, the compact PLaMo 2.1-2B-VL also exceeded the performance of the comparison models. These results indicate that PLaMo 2.1-VL can not only respond naturally in Japanese but also stably generate answers that are grounded in visual content. In particular, the substantial performance gap relative to Qwen3-VL-8B-Instruct (a model in the same size class) suggests that our Japanese-centric training policy and synthetic data pipelines functioned effectively.

\begin{table}[tb]
    \centering
    \caption{VQA evaluation results on JA-VG-VQA-500. Bold indicates proposed models outperforming all previous models.}
    \label{tab:VQA_evaluation_results}
    \small
    \renewcommand{\arraystretch}{1.2}
    \begin{tabularx}{\textwidth}{
        >{\raggedright\arraybackslash}m{0.30\textwidth}
        >{\centering\arraybackslash}X
        >{\centering\arraybackslash}X
        >{\centering\arraybackslash}X
    }
    \toprule
    \multicolumn{1}{c}{\textbf{Model}} &
    \textbf{ROUGE-L} &
    \textbf{LLM-as-a-judge} &
    \textbf{English Likert LLM judge} \\
    \midrule
    \textbf{PLaMo 2.1-8B-VL} & \textbf{61.5} & \textbf{72.4} & \textbf{4.37} \\
    \textbf{PLaMo 2.1-2B-VL} & \textbf{60.7} & \textbf{71.6} & \textbf{4.41} \\
    Asagi-14B & 56.8 & 70.6 & 4.05 \\
    Qwen3-VL-8B-Instruct & 41.6 & 60.4 & 4.06 \\
    \bottomrule
    \end{tabularx}
\end{table}

\subsection{Visual Grounding}
\label{sec:results-vg}
To assess Visual Grounding, we evaluated Referring Expression Comprehension (REC) performance using both the English and Japanese versions of Ref-L4. For our baselines, we selected: (1) Qwen2.5-VL-7B-Instruct~\cite{hf_qwen25_vl_7b_instruct}, which demonstrated the highest performance among similar-sized open models at the start of this project (April 2025); (2) Qwen3-VL-8B-Instruct, the latest open model in the relevant size class at the time of evaluation; and (3) Qwen3-VL-235B-A22B-Instruct~\cite{hf_qwen3_vl_235b_a22b_instruct}, the flagship model of the Qwen3-VL series.

Table~\ref{tab:Visual_Grounding_evaluation_results} shows the results. On both the English and Japanese versions of Ref-L4, PLaMo 2.1-8B-VL outperformed all comparison models. PLaMo 2.1-2B-VL scored slightly lower than Qwen3-VL-8B-Instruct on the English benchmark, but achieved higher scores than the baselines on the Japanese dataset. These results suggest that PLaMo 2.1-VL can identify target regions based on long, complex Japanese instructions. The 8B model performed well in \textit{both} English and Japanese, which suggests that the enhanced grounding capability transfers across the two evaluated languages. The 2B model performed well on the Japanese benchmark, which suggests that our translation and adaptation pipelines are effective even for smaller-parameter models.

\begin{table}[tb]
    \centering
    \caption{Visual Grounding evaluation results on the Japanese and English versions of Ref-L4. Bold indicates proposed models outperforming all previous models.}
    \label{tab:Visual_Grounding_evaluation_results}
    \small
    \begin{tabular}{lrr}
    \toprule
    Model & Japanese Ref-L4 & English Ref-L4 \\
    \midrule
    \textbf{PLaMo 2.1-8B-VL} & \textbf{85.2} & \textbf{86.8} \\
    \textbf{PLaMo 2.1-2B-VL} & \textbf{82.4} & 83.5 \\
    Qwen2.5-VL-7B-Instruct & 76.9 & 83.1 \\
    Qwen3-VL-8B-Instruct & 80.6 & 84.1 \\
    Qwen3-VL-235B-A22B-Instruct & 81.6 & 86.0 \\
    \bottomrule
    \end{tabular}
\end{table}

\subsection{Factory Task Analysis}
\label{sec:results-factory}
For factory task analysis, we conducted zero-shot evaluation on a factory task analysis benchmark. The factory environments and image distributions included in this benchmark were unseen during the training of PLaMo 2.1-VL. We compared our models against Qwen2.5-VL-7B-Instruct, Qwen3-VL-8B-Instruct, and the flagship Qwen3-VL-235B-A22B-Instruct.

Table~\ref{tab:zero-shot_classification_accuracy} shows the results. This setting requires zero-shot classification across ten industrial tasks and is difficult. Even strong open models like Qwen2.5-VL-7B-Instruct (27.6\%), the newer Qwen3-VL-8B-Instruct (38.3\%), and the Qwen3-VL-235B-A22B-Instruct (45.8\%) struggled to achieve high accuracy. In contrast, by synergizing tool-recognition-focused training with Japanese-centric adaptation, PLaMo 2.1-8B-VL achieved 53.9\% accuracy, establishing a clear lead over the baselines. Furthermore, while PLaMo 2.1-2B-VL fell short of the flagship Qwen3-VL-235B-A22B-Instruct, it outperformed Qwen3-VL-8B-Instruct despite having fewer parameters.

\begin{figure}[tb]
\centering
\begin{tabular}{ccc}
    \includegraphics[width=0.3\linewidth]{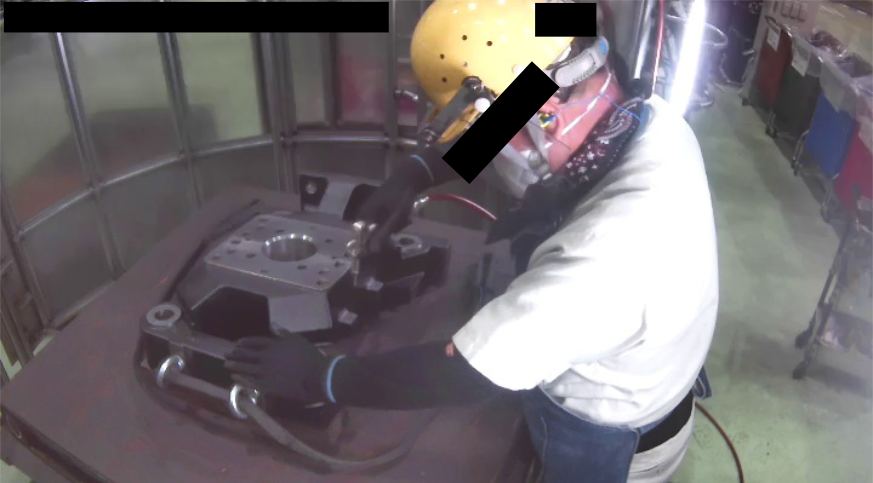} &
    \includegraphics[width=0.3\linewidth]{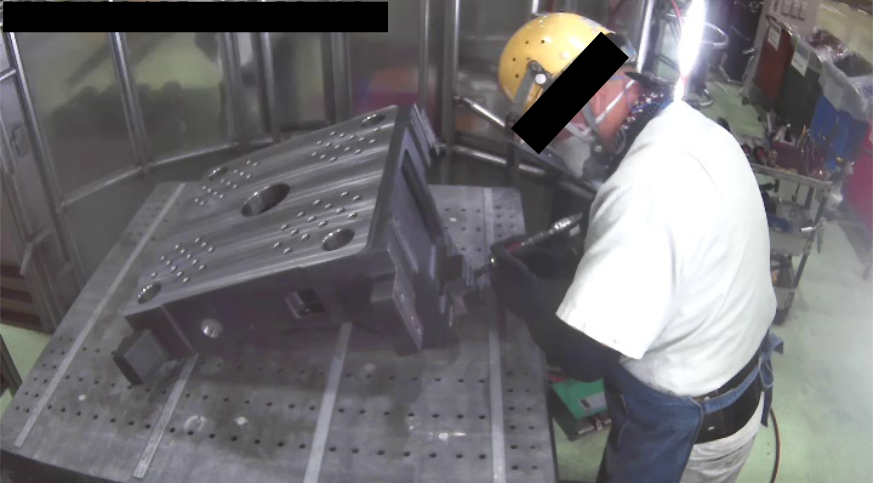} &
    \includegraphics[width=0.3\linewidth]{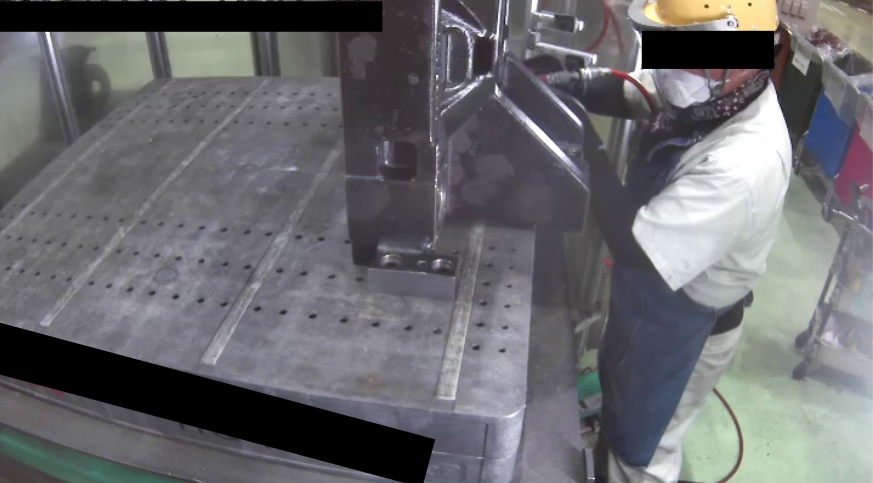} \\
    
    \small \makebox[0pt][r]{Ground truth:} \makebox[0pt][l]{air blow gun} & 
    \small \makebox[0pt][r]{Ground truth:} \makebox[0pt][l]{deburring tool} & 
    \small \makebox[0pt][r]{Ground truth:} \makebox[0pt][l]{\textcolor{red}{deburring tool}} \\
    
    \small \makebox[0pt][r]{Model output:} \makebox[0pt][l]{air blow gun} & 
    \small \makebox[0pt][r]{Model output:} \makebox[0pt][l]{deburring tool} & 
    \small \makebox[0pt][r]{Model output:} \makebox[0pt][l]{air blow gun} \\
\end{tabular}
\caption{PLaMo 2.1-8B-VL prediction examples for factory task analysis. The left and center examples are correct predictions, while the right example is a misclassification.}
\label{fig:examples_factory_task_analysis}
\end{figure}

Figure~\ref{fig:examples_factory_task_analysis} shows example predictions. In this task, many tools share visual similarities, and occlusions by worker hands, bodies, or workpieces lead to misclassifications. In the future, incorporating temporal information (video sequences) alongside single-image frames is expected to mitigate these occlusion issues and further enhance classification accuracy.

\begin{table}[tb]
    \centering
    \caption{Comparison of zero-shot classification accuracy for factory task analysis (10 classes). Bold indicates proposed models outperforming all previous models.}
    \label{tab:zero-shot_classification_accuracy}
    \small
    \begin{tabular}{lr}
    \toprule
    Model & Accuracy \\
    \midrule
    \textbf{PLaMo 2.1-8B-VL} & \textbf{53.9} \\
    \textbf{PLaMo 2.1-2B-VL} & 40.9 \\
    Qwen2.5-VL-7B-Instruct & 27.6 \\
    Qwen3-VL-8B-Instruct & 38.3 \\
    Qwen3-VL-235B-A22B-Instruct & 45.8 \\
    \bottomrule
    \end{tabular}
\end{table}

\subsection{Anomaly Detection for Infrastructure Facilities}
\label{sec:results-anomaly}
This section presents the results on a power plant anomaly detection benchmark. We first report zero-shot performance and analyze common error modes. We then evaluate the effect of domain-specific fine-tuning on real-world operational data.
\subsubsection{Zero-shot}
\label{sec:results-anomaly-zero-shot}
\begin{table}[tb]
    \centering
    \caption{Zero-shot evaluation results on the power plant anomaly detection benchmark. Bold indicates proposed models outperforming all previous models.}
    \label{tab:Zero-shot_evaluation_results}
    \small
    \begin{tabular}{lrr}
    \toprule
    Model & bbox only & bbox + label \\
    \midrule
    \textbf{PLaMo 2.1-8B-VL} & \textbf{58.9} & \textbf{39.3} \\
    \textbf{PLaMo 2.1-2B-VL} & \textbf{57.4} & \textbf{38.0} \\
    Qwen2.5-VL-7B-Instruct & 6.3 & 2.5 \\
    Qwen3-VL-8B-Instruct & 10.7 & 6.1 \\
    Qwen3-VL-235B-A22B-Instruct & 34.5 & 25.1 \\
    \bottomrule
    \end{tabular}
\end{table}

For the zero-shot baseline, we evaluated our models against Qwen2.5-VL-7B-Instruct, Qwen3-VL-8B-Instruct, and Qwen3-VL-235B-A22B-Instruct. In preliminary tests, two-pass inference gave higher label estimation accuracy than single-pass inference for all models. We therefore report all results below using the two-pass inference protocol.

As shown in Table~\ref{tab:Zero-shot_evaluation_results}, both PLaMo 2.1-8B-VL and PLaMo 2.1-2B-VL achieved higher average F1-scores than the comparison models under both operational conditions (bbox only and bbox + label). However, the drop in scores from the bbox-only to the bbox + label condition indicates that the models can localize anomalous regions, but accurate label classification remains a challenge.

\begin{figure}[tb]
    \centering
    \begin{tabular}{cc}
    
        \footnotesize \makebox[0pt][r]{Target image (abnormal)} \quad \makebox[0pt][l]{Reference image (normal)} & 
        \footnotesize \makebox[0pt][r]{Target image (abnormal)} \quad \makebox[0pt][l]{Reference image (normal)} \\
    
        \includegraphics[width=0.49\linewidth]{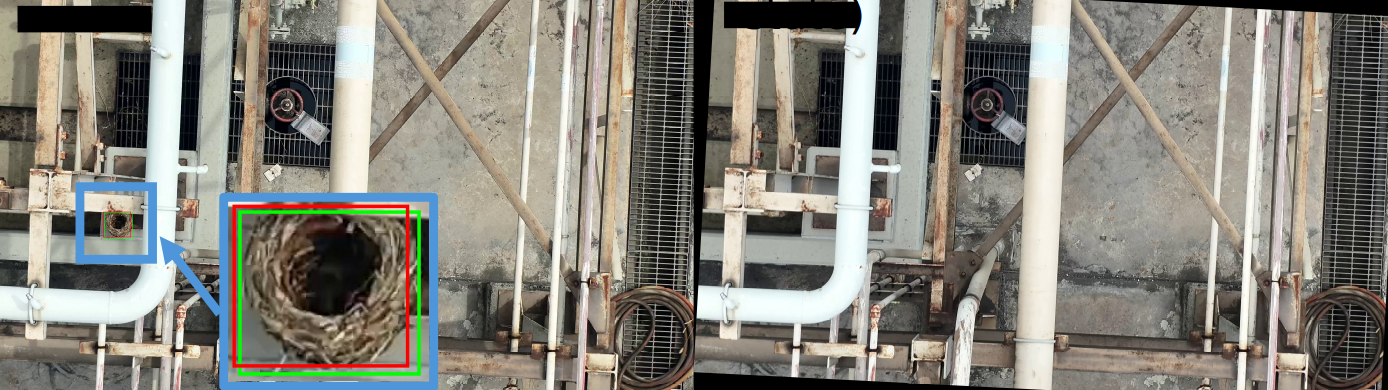} &
        \includegraphics[width=0.49\linewidth]{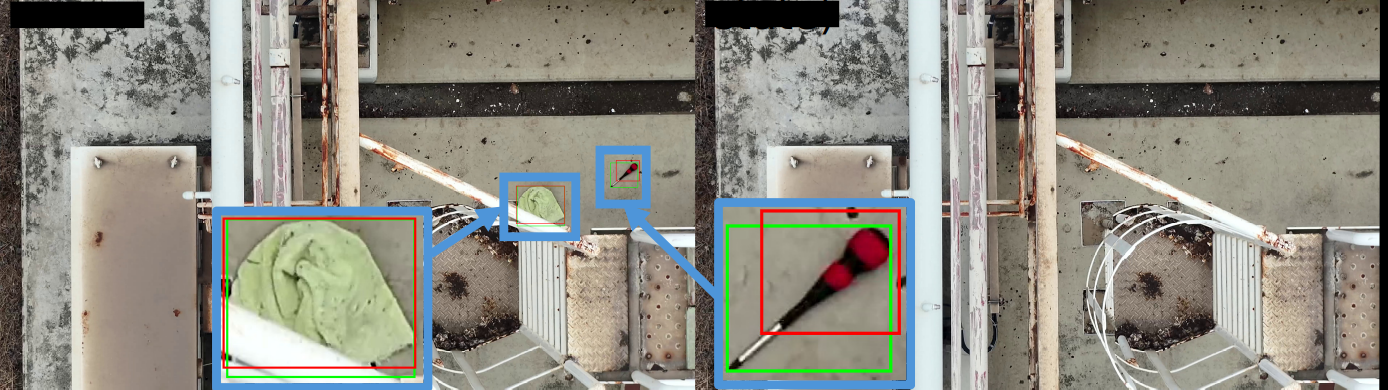} \\
    
        \small \makebox[0pt][r]{\textcolor{promptgreen}{Green}: Ground truth} \quad \makebox[0pt][l]{Ground truth: bird nest} & 
        \small \makebox[0pt][r]{Ground truth: towel} \quad \makebox[0pt][l]{Ground truth: tool} \\
    
        \small \makebox[0pt][r]{\phantom{Gr}\textcolor{red}{Red}: Model output} \quad \makebox[0pt][l]{Model output: bird nest} & 
        \small \makebox[0pt][r]{Model output: towel} \quad \makebox[0pt][l]{Model output: tool} \\
    
        \includegraphics[width=0.49\linewidth]{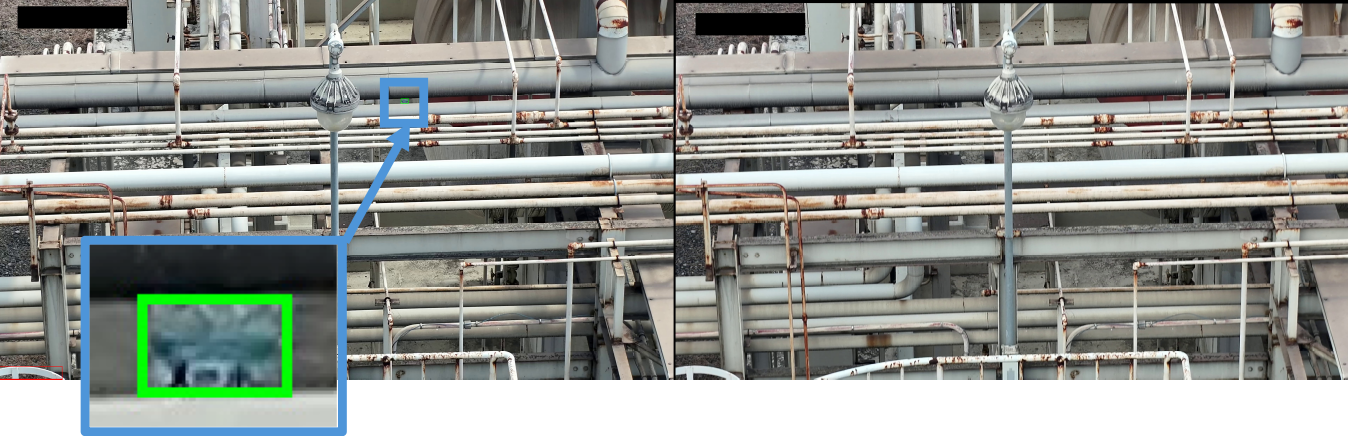} &
        \includegraphics[width=0.49\linewidth]{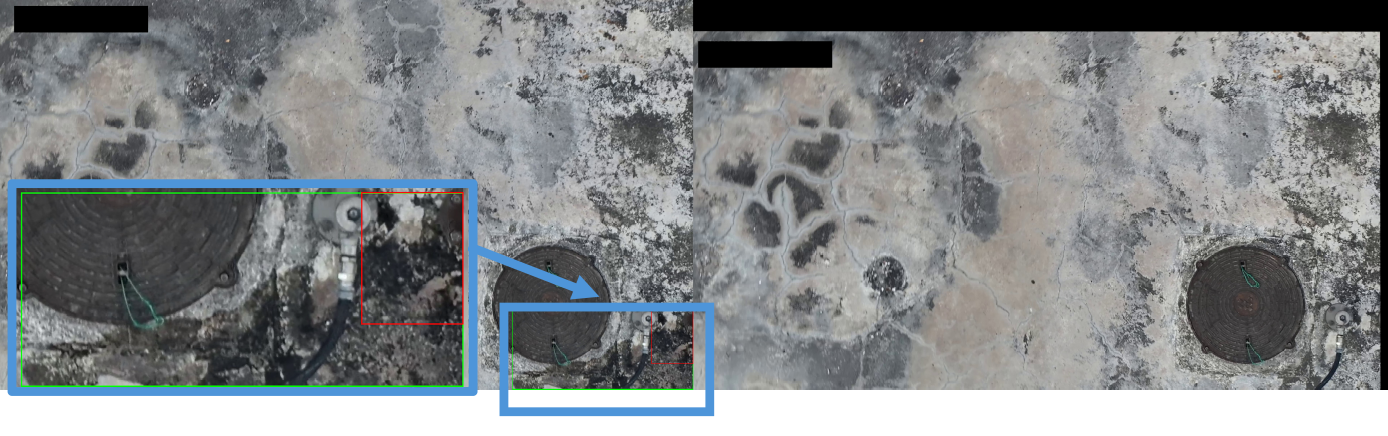} \\
        
        \small \makebox[0pt][r]{Ground truth:} \makebox[0pt][l]{plastic bottle} & 
        \small \makebox[0pt][r]{Ground truth:} \makebox[0pt][l]{water leakage} \\
    
        \small \makebox[0pt][r]{Model output:} \makebox[0pt][l]{$\langle$not detected$\rangle$} & 
        \small \makebox[0pt][r]{Model output:} \makebox[0pt][l]{person} \\
    \end{tabular}
    \caption{Prediction examples of anomaly detection by PLaMo 2.1-8B-VL. The upper row shows successful cases, and the lower row shows failures. The failures illustrate an example of missing a small object and an example of mismatch in bounding box and label for a state-driven anomaly.}
    \label{fig:anomaly_detection}
\end{figure}

Figure~\ref{fig:anomaly_detection} shows example predictions. In the top row, the model correctly localizes and labels object anomalies, such as bird nests, towels, and tools. The bottom row shows common error modes: missing visually small anomalies (e.g., a plastic bottle) and mislabeling state-driven anomalies (e.g., water leakage).

\begin{figure}[htb!]
    \centering
    \includegraphics[width=0.57\linewidth]{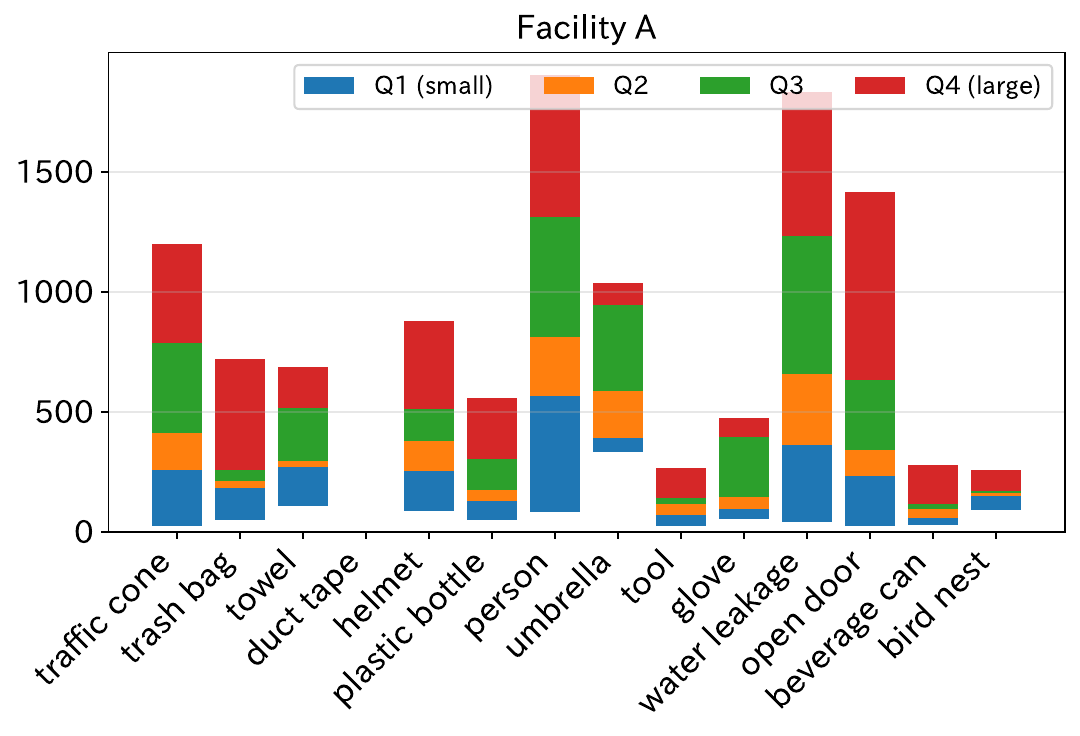}
    \includegraphics[width=0.57\linewidth]{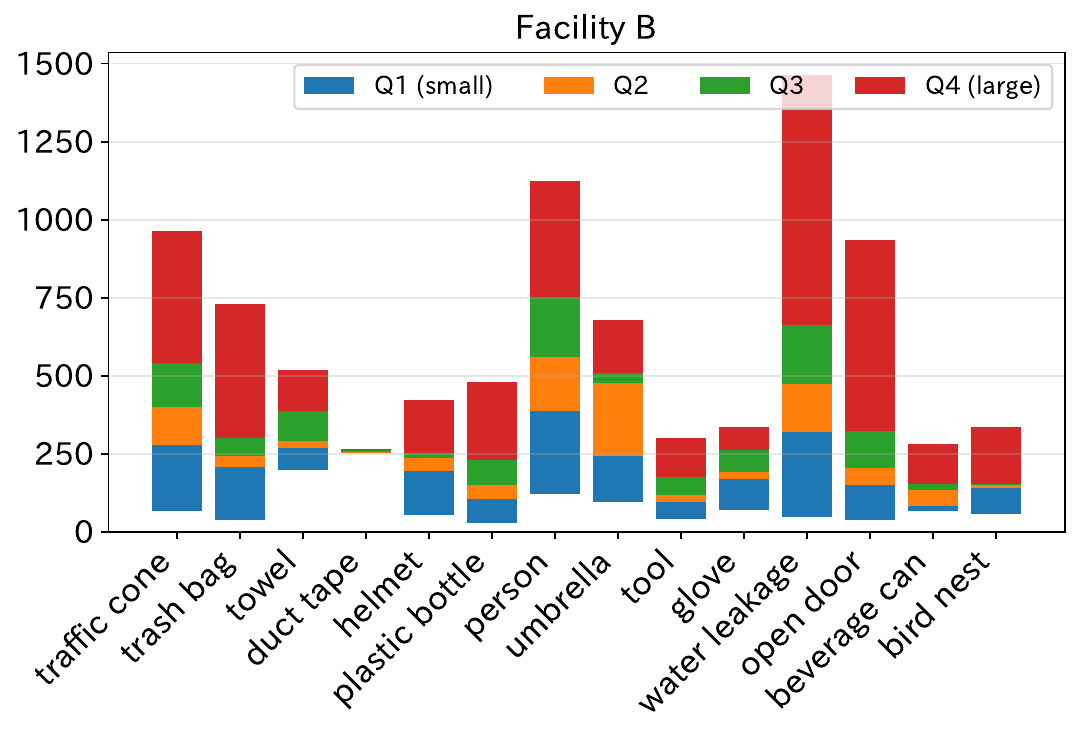}
    \includegraphics[width=0.57\linewidth]{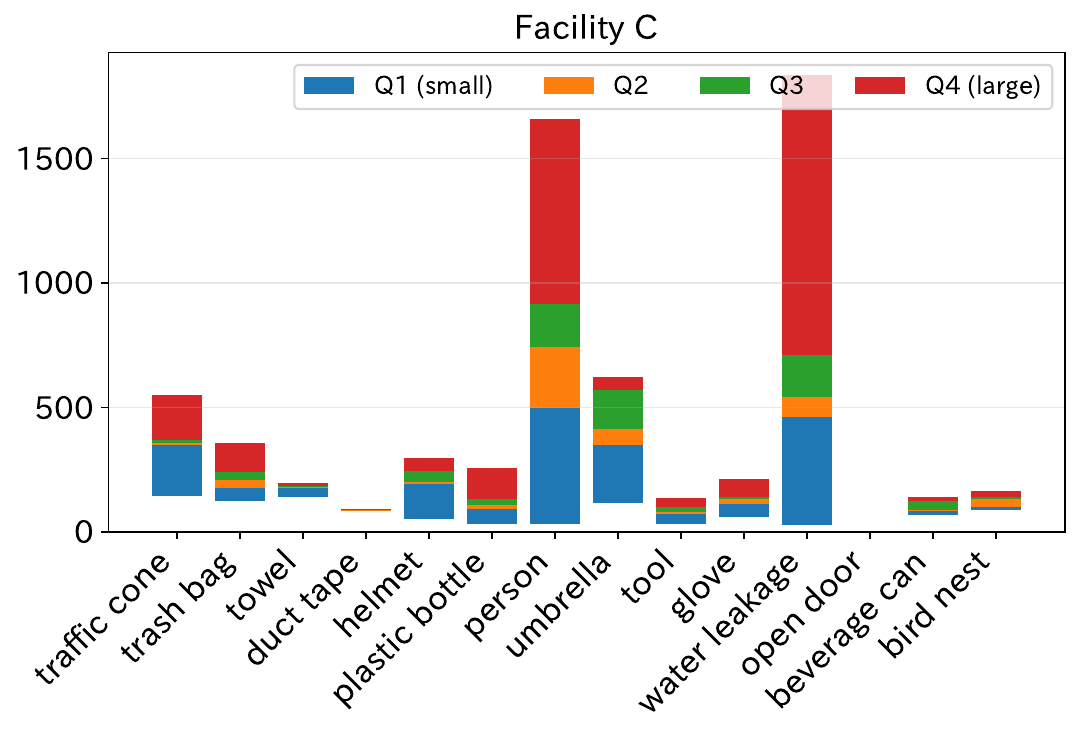}
    \caption{Distribution of bounding box sizes in the constructed dataset. Bounding boxes are sorted in ascending order of size and divided into four quantiles (Q1 to Q4) with equal numbers of samples (Q1 smallest, Q4 largest), and the size range for each quantile is shown. The vertical axis is the geometric mean of height and width of the bounding box (that is, the square root of the area).}
    \label{fig:distribution_bbox_sizes}
\end{figure}

\begin{figure}[htb!]
    \centering
    \includegraphics[width=0.55\linewidth]{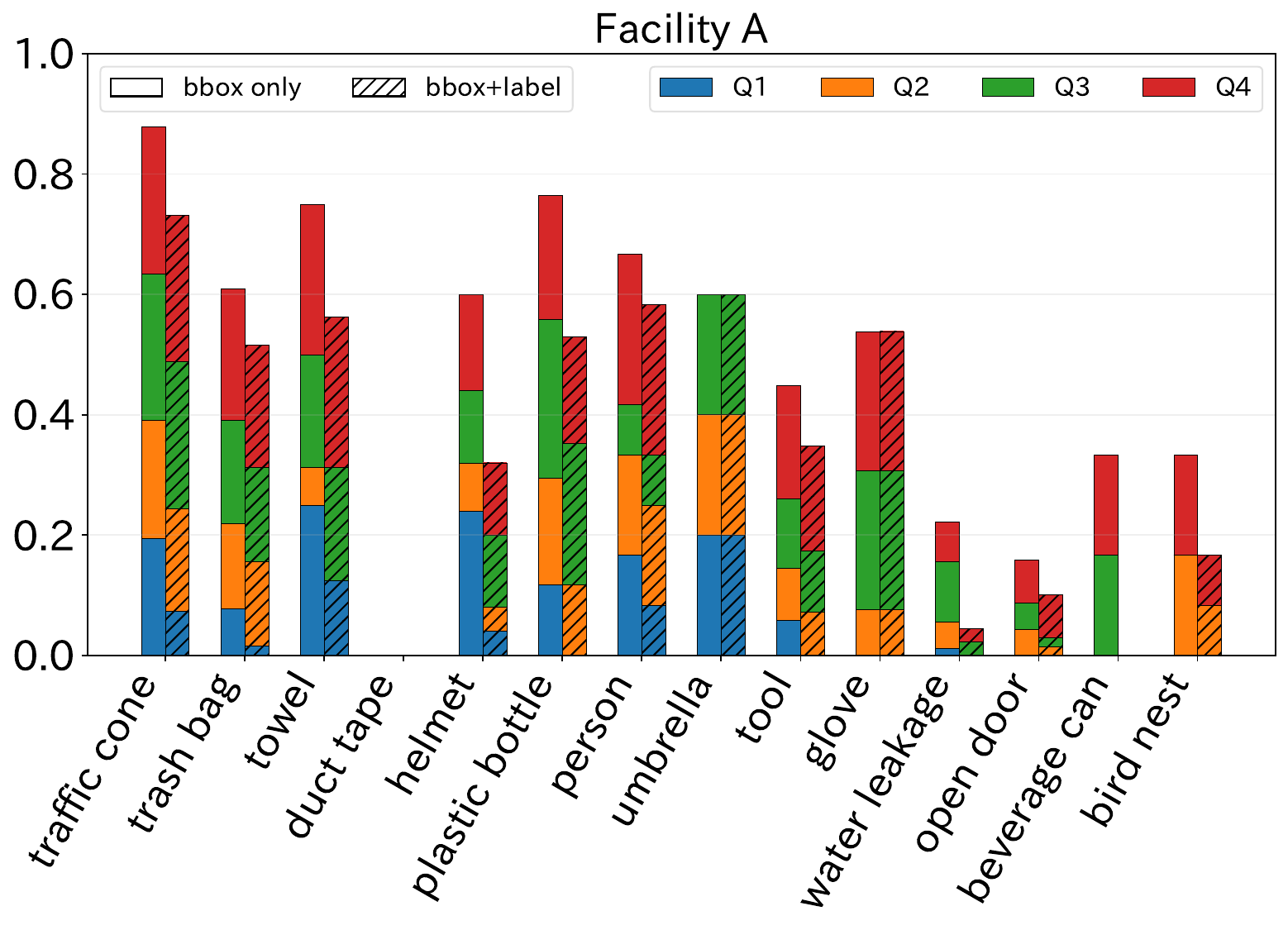}
    \includegraphics[width=0.55\linewidth]{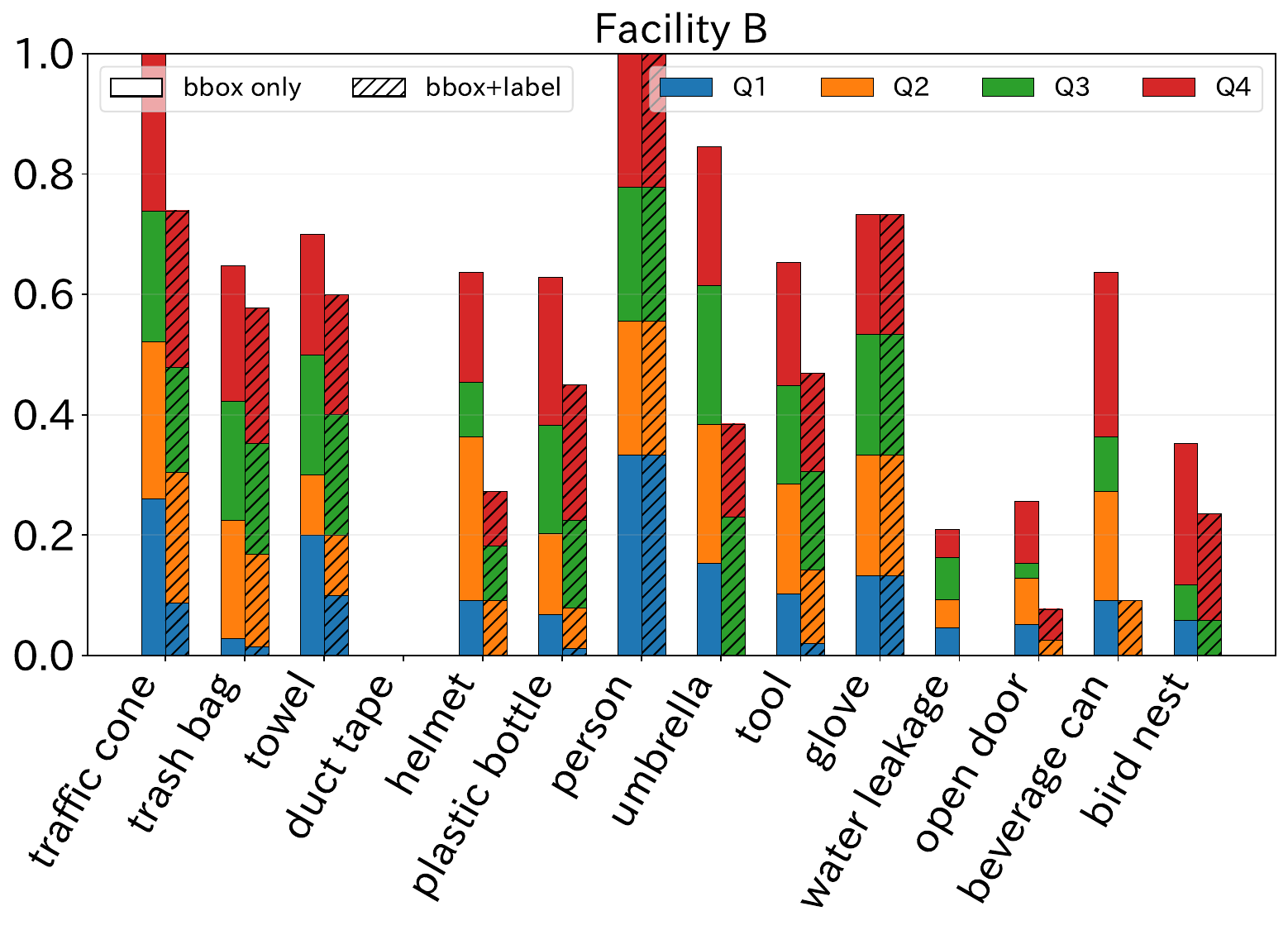}
    \includegraphics[width=0.55\linewidth]{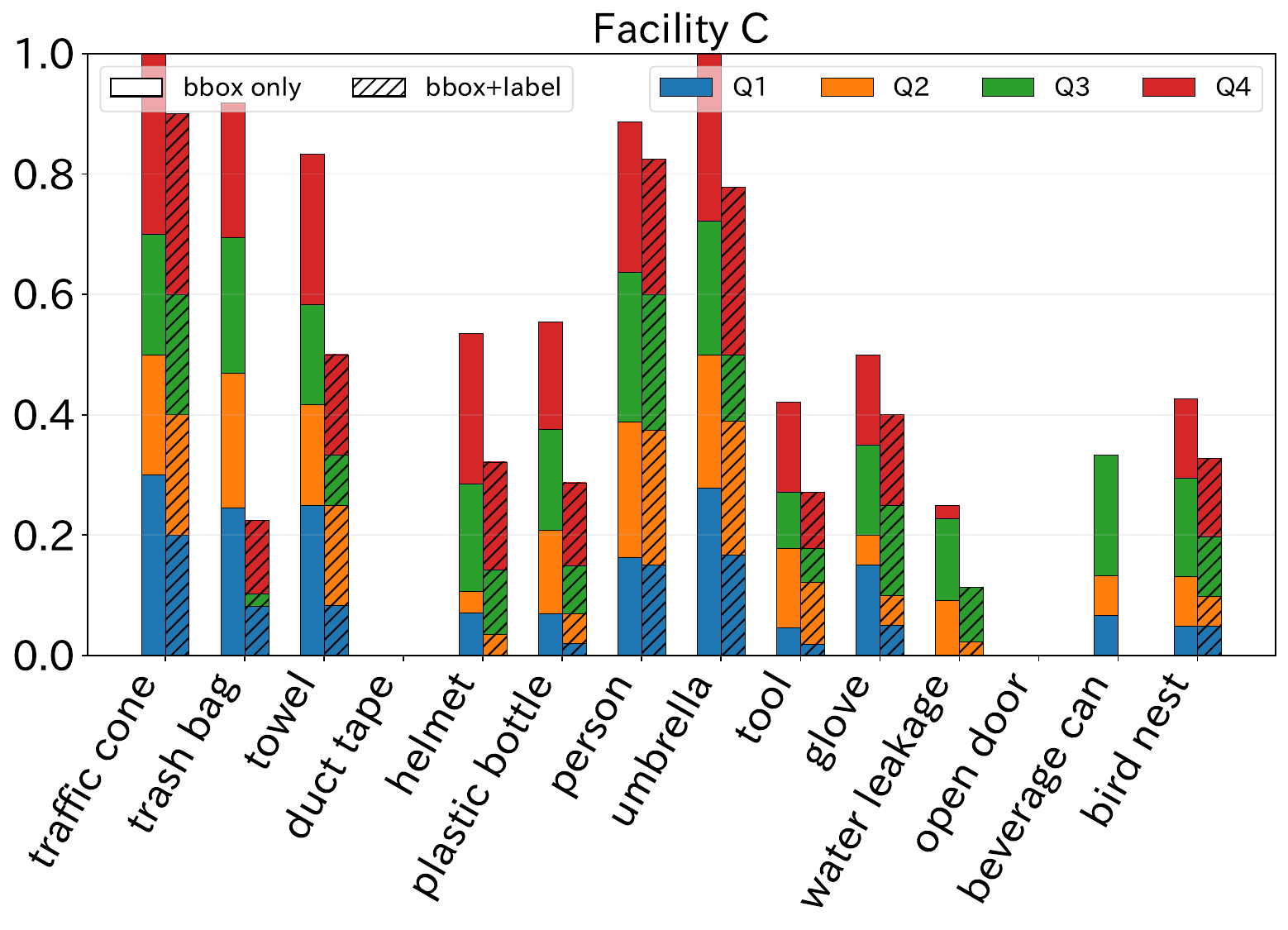}
    \caption{Breakdown of agreement rates in PLaMo 2.1-8B-VL's detection results. For each bounding box size quantile (Q1 to Q4), this figure shows the proportion of bounding boxes that matched under the bbox-only and bbox + label conditions. The maximum agreement rate in each quantile is 0.25, and the sum of agreement rates reaches 1.0 if all bounding boxes in all four quantiles match.}
    \label{fig:breakdown_agreement_rates}
\end{figure}

To analyze error patterns in PLaMo 2.1-8B-VL, Figure~\ref{fig:distribution_bbox_sizes} shows the distribution of bounding box sizes for anomalous objects, and Figure~\ref{fig:breakdown_agreement_rates} shows the agreement rates by bounding box size and anomaly type. In both figures, bounding boxes are sorted by size and divided into four equal-sample quartiles (Q1 to Q4, with Q1 being the smallest). Two trends appear:

First, agreement rates increase with bounding box size (Q4 \textgreater{} Q3 \textgreater{} Q2 \textgreater{} Q1). Small-object anomalies, such as beverage cans and plastic bottles, are difficult to both detect and label, and their agreement rates remain low. The missed detection in the lower-left of Figure~\ref{fig:anomaly_detection} is an example of this.

Second, state-driven anomalies, such as open doors or liquid leakage, often cause errors in both spatial localization and semantic labeling, especially in the bbox + label metric. In the failure case shown in the lower-right of Figure~\ref{fig:anomaly_detection}, the model predicts a region near the actual leakage, but the IoU falls below 0.5, so it counts as an error for both the box and the label. This suggests that, for state-driven anomalies, both finding the exact boundary and assigning the correct label are difficult. From a deployment perspective, these zero-shot results suggest that workflows relying on ``visual confirmation triggered by localized alerts'' (bbox only) are easier to deploy, whereas fully automated reporting systems (bbox + label) require further improvement.

To isolate the effect of appearance conditions relevant to operational decision-making, specifically target size and shooting distance, we removed the state-driven anomalies (299 samples) and reevaluated the dataset. We progressively imposed a lower-bound threshold on the bounding box size (specifically, the square root of the area), excluding samples that fell below the threshold (Table~\ref{tab:Anomaly_detection_results}). As expected, increasing the size threshold (thereby restricting the evaluation to visually prominent anomalies) increased the average F1-scores. These results suggest that to achieve reliable performance in production environments, drone camera specifications and flight parameters (altitude and optical zoom) must be calibrated to ensure the target anomalies meet a minimum pixel-size threshold. For reference, with the Zenmuse H30 zoom camera used for this development (video $3840 \times 2160$, diagonal field of view (DFOV) $66.7$\textdegree), we estimate that a standard $350~\mathrm{\,mL}$ can (height $\sim0.12~\mathrm{\,m}$) requires a subject distance of approximately 12 to $20~\mathrm{\,m}$ at 3x to 5x zoom to register a geometric mean of 100 pixels on-screen, assuming motion blur during flight is kept minimal.

\begin{table}[tb]
    \centering
    \caption{Anomaly detection results after excluding state-driven anomalies and imposing lower bounds on bbox size.}
    \label{tab:Anomaly_detection_results}
    \small
    \begin{tabularx}{\textwidth}{>{\raggedright\arraybackslash}X>{\raggedleft\arraybackslash}X>{\raggedleft\arraybackslash}X>{\raggedleft\arraybackslash}X}
    \toprule
    bbox size (geometric mean) & Number of samples & bbox only & bbox + label \\
    \midrule
    0 px or larger & 901 / 1200 & 68.4 & 48.5 \\
    100 px or larger & 679 / 1200 & 76.8 & 56.5 \\
    200 px or larger & 360 / 1200 & 88.0 & 72.8 \\
    \bottomrule
    \end{tabularx}
\end{table}

\subsubsection{Fine-tuning on Plant Data}
\label{sec:results-anomaly-finetuning}
\begin{table}[t]
    \centering
    \caption{Evaluation results of fine-tuning PLaMo 2.1-8B-VL using power plant data (the evaluation set differs from that of Table~\ref{tab:Zero-shot_evaluation_results}).}
    \label{tab:fine-tuning}
    \small
    \begin{tabular}{lrr}
    \toprule
    PLaMo 2.1-8B-VL model & bbox only & bbox + label \\
    \midrule
    Base model & 65.9 & 39.7 \\
    Fine-tuning & 71.6 & 64.9 \\
    \bottomrule
    \end{tabular}
\end{table}

After establishing the zero-shot baselines, we next evaluated the effect of fine-tuning PLaMo 2.1-8B-VL using real-world data collected at a power plant for this target domain.

For fine-tuning, we used data collected over three days at three plant facilities (A, B, and C). Data from days 1 to 3 at facilities A and B were used as the training set (54,722 samples). Data from day 1 at facility C were used as the validation set (5,172 samples). Finally, 2,000 anomaly-containing samples were randomly selected from days 2 and 3 at facility C as an unseen evaluation set. Each sample consisted of a target/reference image pair with bounding boxes and labels for the anomalies.

To match training to the inference pipeline, we processed the training and validation data using the same two-pass architecture. In pass 1, the model was trained to predict bounding boxes and labels from the full target/reference composite image. In pass 2, cropped regions containing the potential anomalies were extracted from the target image and used as localized inputs to re-estimate the labels. We used early stopping based on validation loss, which triggered after consuming roughly 20\% of the full training data.

Table~\ref{tab:fine-tuning} compares the base model and the fine-tuned model on this evaluation set. Fine-tuning improved performance of PLaMo 2.1-8B-VL in both pass 1 and pass 2. Notably, the improvement was far more pronounced in the bbox + label evaluation compared to the bbox-only evaluation. This suggests that while spatial localization capabilities were already accurate in the zero-shot setting, semantic labeling improved more after adaptation.

We further evaluated this fine-tuned model by imposing the previously discussed bounding box size thresholds (Table~\ref{tab:fine-tuned_model_after_excluding}). The results show that label recognition for small anomalies (under 200 px) improved. 
As in the zero-shot setting, these results suggest that, for real-world deployment, camera specifications and imaging conditions should be chosen based on the size of the anomalies to be detected.

Operationally, capturing and annotating data inside an active power plant is costly, and artificially staging physical anomalies inside active facilities introduces severe operational constraints. However, the results suggest a practical direction: because spatial localization (bounding box prediction) is already accurate, the primary remaining bottleneck is label estimation. Given that pass 2 requires only the cropped region to re-estimate the label, and is therefore less affected by complex industrial backgrounds, it may not be necessary to limit data collection exclusively to the target facility. Future efforts to improve semantic labeling could leverage cropped anomaly candidates gathered from other low-risk environments.

\begin{table}[tb]
    \centering
    \caption{Evaluation results of the base model (PLaMo 2.1-8B-VL) and the fine-tuned model after excluding state-driven anomalies and imposing lower bounds on bbox area.}
    \label{tab:fine-tuned_model_after_excluding}
    \small
    \begin{tabularx}{\textwidth}{>{\raggedright\arraybackslash}X>{\raggedleft\arraybackslash}X>{\raggedleft\arraybackslash}X>{\raggedleft\arraybackslash}X>{\raggedleft\arraybackslash}X>{\raggedleft\arraybackslash}X}
    \toprule
    bbox size (geometric mean) & Number of samples & Base model bbox only & Base model bbox + label & Fine-tuning bbox only & Fine-tuning bbox + label \\
    \midrule
    {\footnotesize 0 px or larger} & 1882 / 2000 & 68.6 & 40.9 & 73.8 & 66.9 \\
    {\footnotesize 100 px or larger} & 493 / 2000 & 96.8 & 69.8 & 97.6 & 90.7 \\
    {\footnotesize 200 px or larger} & 151 / 2000 & 98.7 & 96.7 & 96.2 & 96.2 \\
    \bottomrule
    \end{tabularx}
\end{table}

\section{Limitations}
\label{sec:limitations}
PLaMo 2.1-VL is a family of models trained mainly for VQA and Visual Grounding on natural images. Therefore, high performance is not necessarily guaranteed on tasks outside this scope. For example, these models have not been trained specifically for OCR, and their ability to read text in images and understand document images, charts, and mathematical formulas is limited. In addition, these models accept only a single image as input. While these models have a certain amount of general and everyday knowledge, they have not been trained intensively on domain-specific expert knowledge. When applying these models to tasks that require expert judgment, additional training, access to external knowledge, or use in combination with surrounding systems may be required.

Furthermore, these models are trained for paired image-text input. Therefore, when they are used with image-only or text-only input, they may not behave as intended. They are also not designed for settings in which both image and text are provided, but the prompt instructs the model not to refer to the image.
\section{Conclusion and Future Outlook}
\label{sec:conclusion}
In this report, we introduced PLaMo 2.1-VL, a lightweight VLM designed for autonomous devices. We evaluated the models on their core capabilities, including VQA and Visual Grounding, as well as on real-world application tasks such as factory task analysis and infrastructure anomaly detection. On standard benchmarks, PLaMo 2.1-8B-VL outperformed open models of similar size, while PLaMo 2.1-2B-VL also performed well on Japanese-language benchmarks despite its smaller size. These results suggest that our model design, training policy, and data construction and translation pipelines, all emphasizing Japanese-language operation, were effective.

However, in real-world deployment, factors such as variable camera conditions, site-specific objects and procedures, and edge cases affect performance; thus, high benchmark scores do not always translate into operational reliability. Consequently, broad zero-shot application of the current PLaMo 2.1-VL still faces challenges. To bridge this gap, system design that incorporates application-specific fine-tuning and clear operational conditions remains crucial.

Nevertheless, this report demonstrates that PLaMo 2.1-VL has potential for practical deployment. For example, in anomaly detection, performance improves when conditions such as target size and shooting distance are appropriately controlled. Furthermore, fine-tuning using field data leads to operational improvements, particularly in label estimation. This suggests that real-world deployment is feasible if the operational scope is properly identified, task settings are constrained to acceptable real-world conditions, and additional training is applied as needed.

Going forward, we will continue to expand the scale and improve the quality of our training data to elevate the model's core capabilities. Our deployment strategy will proceed in two stages. First, we will identify limited use cases where the current PLaMo 2.1-VL can provide value, rolling it out gradually through proof-of-concept deployments. These limited use cases involve domains where the target scope and decision criteria are clear, assumptions regarding shooting conditions and target sizes are well-defined, and recovery strategies (such as human-in-the-loop confirmation or rule-based fallbacks) can be incorporated. In such environments, operational value can be realized early, even in a zero-shot setting or with minimal adaptation. Throughout this phase, operational logs and failure cases will be continuously collected to drive an iterative improvement cycle.

Second, for use cases that involve numerous exceptions, require the detection of subtle differences, or carry high costs for misjudgment, we will collect targeted field data and utilize fine-tuning to adapt the model to specific requirements. While this process may require effort comparable to conventional machine learning pipelines, VLMs retain the advantage of strong generalization, allowing pretrained capabilities to be reused across a broad range of settings. Furthermore, because decision-making criteria can be managed via natural language, VLMs offer the advantage of being able to respond flexibly to operational changes and task extensions.

Ultimately, by pursuing a parallel strategy of early deployment in constrained use cases, careful design of operational conditions, and targeted fine-tuning based on field data, we will overcome the limitations of zero-shot inference and steadily advance toward robust systems capable of reliable, real-world operation.
\section{Contributions and Acknowledgements}
\label{sec:Contributions_acknowledgement}
\subsection{Authors}
All contributors of PLaMo 2.1-VL are listed in alphabetical order by their last names.

\vspace{\baselineskip}
\noindent Tommi Kerola, Yuya Masuda, Takashi Masuko, Toshiki Nakanishi, Daisuke Nishino, Kuniyuki Takahashi, Hanqin Wang, Yoshihiro Yamada

\subsection{Acknowledgements}
\label{sec:acknowledgement}
This model was created under the ``GENIAC (Generative AI Accelerator Challenge)'' project, implemented by the Ministry of Economy, Trade and Industry (METI) and the New Energy and Industrial Technology Development Organization (NEDO), with the aim of strengthening Japan's development capabilities in generative AI.
\bibliographystyle{unsrt}
\bibliography{reference}

@article{liu2023visual,
  title={Visual instruction tuning},
  author={Liu, Haotian and Li, Chunyuan and Wu, Qingyang and Lee, Yong Jae},
  journal={{Advances in Neural Information Processing Systems}},
  volume={36},
  pages={34892--34916},
  year={2023}
}

@misc{networks2025plamo,
  title={{PLaMo 2 Technical Report}},
  author={{Preferred Networks, Inc.} and Chubachi, Kaizaburo and Fujita, Yasuhiro and Hemmi, Shinichi and Hirokawa, Yuta and Imajo, Kentaro and Kataoka, Toshiki and Kobayashi, Goro and Maehashi, Kenichi and Metzger, Calvin and others},
  howpublished = {\url{https://arxiv.org/abs/2509.04897}},
  note         = {arXiv preprint arXiv:2509.04897},
  year={2025}
}

@misc{tschannen2025siglip,
  author       = {Tschannen, Michael and Gritsenko, Alexey and Wang, Xiao and Naeem, Muhammad Ferjad and Alabdulmohsin, Ibrahim and Parthasarathy, Nikhil and Evans, Talfan and Beyer, Lucas and Xia, Ye and Mustafa, Basil and others},
  title        = {{SigLIP 2: Multilingual Vision-Language Encoders with Improved Semantic Understanding, Localization, and Dense Features}},
  year         = {2025},
  howpublished = {\url{https://arxiv.org/abs/2502.14786}},
  note         = {arXiv preprint arXiv:2502.14786}
}

@misc{hf_siglip2_so400m_patch14_384,
  author       = {{Google DeepMind}},
  title        = {{google/siglip2-so400m-patch14-384}},
  howpublished = {\url{https://huggingface.co/google/siglip2-so400m-patch14-384}},
  note         = {Accessed: 2026-04-08}
}

@misc{li2025eagle,
  title={{Eagle 2: Building Post-Training Data Strategies from Scratch for Frontier Vision-Language Models}},
  author={Li, Zhiqi and Chen, Guo and Liu, Shilong and Wang, Shihao and VS, Vibashan and Ji, Yishen and Lan, Shiyi and Zhang, Hao and Zhao, Yilin and Radhakrishnan, Subhashree and others},
  year         = {2025},
  howpublished = {\url{https://arxiv.org/abs/2501.14818}},
  note         = {arXiv preprint arXiv:2501.14818}
}

@misc{beyer2024paligemma,
  title={{PaliGemma: A versatile 3B VLM for transfer}},
  author={Beyer, Lucas and Steiner, Andreas and Pinto, Andr{\'e} Susano and Kolesnikov, Alexander and Wang, Xiao and Salz, Daniel and Neumann, Maxim and Alabdulmohsin, Ibrahim and Tschannen, Michael and Bugliarello, Emanuele and others},
  year         = {2024},
  howpublished = {\url{https://arxiv.org/abs/2407.07726}},
  note         = {arXiv preprint arXiv:2407.07726}
}

@misc{qwen25_vl_2025,
    author={Bai, Shuai and Chen, Keqin and Liu, Xuejing and Wang, Jialin and Ge, Wenbin and Song, Sibo and Dang, Kai and Wang, Peng and Wang, Shijie and Tang, Jun and Zhong, Humen and Zhu, Yuanzhi and Yang, Mingkun and Li, Zhaohai and Wan, Jianqiang and Wang, Pengfei and Ding, Wei and Fu, Zheren and Xu, Yiheng and Ye, Jiabo and Zhang, Xi and Xie, Tianbao and Cheng, Zesen and Zhang, Hang and Yang, Zhibo and Xu, Haiyang and Lin, Junyang},
    title        = {{Qwen2.5-VL Technical Report}},
    year         = {2025},
    howpublished = {\url{https://arxiv.org/abs/2502.13923}},
    note         = {arXiv preprint arXiv:2502.13923}
}

@misc{bai2025qwen3,
  title={{Qwen3-VL Technical Report}},
  author={Bai, Shuai and Cai, Yuxuan and Chen, Ruizhe and Chen, Keqin and Chen, Xionghui and Cheng, Zesen and Deng, Lianghao and Ding, Wei and Gao, Chang and Ge, Chunjiang and others},
      year         = {2025},
      howpublished = {\url{https://arxiv.org/abs/2511.21631}},
      note         = {arXiv preprint arXiv:2511.21631}
  
}

@inproceedings{li2023blip,
  author={Li, Junnan and Li, Dongxu and Savarese, Silvio and Hoi, Steven},
  title        = {{BLIP-2: Bootstrapping Language-Image Pre-training with Frozen Image Encoders and Large Language Models}},
  booktitle={{International Conference on Machine Learning}},
  pages={19730--19742},
  year={2023},
  organization={PMLR}
}

@article{navit_2023,
  title        = {{Patch n' Pack: NaViT, a Vision Transformer for any Aspect Ratio and Resolution}},
  author={Dehghani, Mostafa and Mustafa, Basil and Djolonga, Josip and Heek, Jonathan and Minderer, Matthias and Caron, Mathilde and Steiner, Andreas and Puigcerver, Joan and Geirhos, Robert and Alabdulmohsin, Ibrahim M and others},
  journal={{Advances in Neural Information Processing Systems}},
  volume={36},
  pages={2252--2274},
  year={2023}
}

@misc{hf_ja_vg_vqa_500,
  author       = {{Sakana AI}},
  title        = {{SakanaAI/JA-VG-VQA-500}},
  howpublished = {\url{https://huggingface.co/datasets/SakanaAI/JA-VG-VQA-500}},
  note         = {Accessed: 2026-04-08}
}

@misc{hf_ref_l4,
  author={Chen, Jierun and Wei, Fangyun and Zhao, Jinjing and Song, Sizhe and Wu, Bohuai and Peng, Zhuoxuan and Chan, S-H Gary and Zhang, Hongyang},
  title        = {{JierunChen/Ref-L4}},
  howpublished = {\url{https://huggingface.co/datasets/JierunChen/Ref-L4}},
  note         = {Accessed: 2026-04-08}
}

@misc{github_ja_ref_l4,
  author       = {{Preferred Networks, Inc.}},
  title        = {{Ja-Ref-L4}},
  year         = {2025},
  howpublished = {\url{https://github.com/pfnet-research/Ja-Ref-L4}},
  note         = {Accessed: 2026-04-08}
}

@inproceedings{yu2016modeling,
  title={{Modeling Context in Referring Expressions}},
  author={Yu, Licheng and Poirson, Patrick and Yang, Shan and Berg, Alexander C and Berg, Tamara L},
  booktitle={{European Conference on Computer Vision}},
  pages={69--85},
  year={2016},
  organization={Springer}
}

@misc{carion2025sam3segmentconcepts,
      title={{SAM 3: Segment Anything with Concepts}},
      author={Nicolas Carion and Laura Gustafson and Yuan-Ting Hu and Shoubhik Debnath and Ronghang Hu and Didac Suris and Chaitanya Ryali and Kalyan Vasudev Alwala and Haitham Khedr and Andrew Huang and Jie Lei and Tengyu Ma and Baishan Guo and Arpit Kalla and Markus Marks and Joseph Greer and Meng Wang and Peize Sun and Roman Rädle and Triantafyllos Afouras and Effrosyni Mavroudi and Katherine Xu and Tsung-Han Wu and Yu Zhou and Liliane Momeni and Rishi Hazra and Shuangrui Ding and Sagar Vaze and Francois Porcher and Feng Li and Siyuan Li and Aishwarya Kamath and Ho Kei Cheng and Piotr Dollár and Nikhila Ravi and Kate Saenko and Pengchuan Zhang and Christoph Feichtenhofer},
      year         = {2025},
      howpublished = {\url{https://arxiv.org/abs/2511.16719}},
      note         = {arXiv preprint arXiv:2511.16719}
}

@misc{hf_qwen25_vl_32b_instruct,
  author       = {{Qwen Team}},
  title        = {{Qwen/Qwen2.5-VL-32B-Instruct}},
  year         = {2025},
  howpublished = {\url{https://huggingface.co/Qwen/Qwen2.5-VL-32B-Instruct}},
  note         = {Accessed: 2026-04-08}
}

@misc{hf_qwen3_32b,
  author       = {{Qwen Team}},
  title        = {{Qwen/Qwen3-32B}},
  year         = {2025},
  howpublished = {\url{https://huggingface.co/Qwen/Qwen3-32B}},
  note         = {Accessed: 2026-04-08}
}

@misc{openimages_v7,
  author       = {{Open Images}},
  title        = {{Open Images Dataset V7}},
  howpublished = {\url{https://storage.googleapis.com/openimages/web/index.html}},
  note         = {Accessed: 2026-04-08}
}

@misc{flickr_p1210645a,
  author       = {Tom May},
  title        = {{P1210645-a}},
  year         = {2015},
  howpublished = {\url{https://www.flickr.com/photos/sleepyhammer/16541842339}},
  note         = {Accessed: 2026-04-08}
}

@misc{cc_by_20,
  author       = {{Creative Commons}},
  title        = {{Attribution 2.0 Generic (CC BY 2.0)}},
  howpublished = {\url{https://creativecommons.org/licenses/by/2.0/}},
  note         = {Accessed: 2026-04-08}
}

@misc{flickr_sunway_lagoon,
  author       = {Mohd Fazlin Mohd Effendy Ooi},
  title        = {{Sunway Lagoon}},
  year         = {2015},
  howpublished = {\url{https://www.flickr.com/photos/phalinn/21104786682}},
  note         = {Accessed: 2026-04-08}
}

@misc{hf_plamo2_translate,
  author       = {{Preferred Networks, Inc.}},
  title        = {{pfnet/plamo-2-translate}},
  year         = {2025},
  howpublished = {\url{https://huggingface.co/pfnet/plamo-2-translate}},
  note         = {Accessed: 2026-04-08}
}

@misc{hf_plamo_embedding,
  author       = {{Preferred Networks, Inc.}},
  title        = {{pfnet/plamo-embedding-1b}},
  year         = {2025},
  howpublished = {\url{https://huggingface.co/pfnet/plamo-embedding-1b}},
  note         = {Accessed: 2026-04-08}
}

@misc{pexels_license,
  author       = {{Pexels}},
  title        = {{Pexels License}},
  howpublished = {\url{https://www.pexels.com/ja-JP/license/}},
  note         = {Accessed: 2026-04-08}
}

@misc{pexels_2821220,
  author       = {{Pexels}},
  title        = {{Photo 2821220}},
  howpublished = {\url{https://www.pexels.com/ja-jp/photo/2821220/}},
  note         = {Accessed: 2026-04-08}
}

@misc{hf_asagi_14b,
  author       = {{MIL-UT}},
  title        = {{MIL-UT/Asagi-14B}},
  howpublished = {\url{https://huggingface.co/MIL-UT/Asagi-14B}},
  note         = {Accessed: 2026-04-08}
}

@misc{hf_qwen3_vl_8b_instruct,
  author       = {{Qwen Team}},
  title        = {{Qwen/Qwen3-VL-8B-Instruct}},
  year         = {2025},
  howpublished = {\url{https://huggingface.co/Qwen/Qwen3-VL-8B-Instruct}},
  note         = {Accessed: 2026-04-08}
}

@misc{hf_qwen25_vl_7b_instruct,
  author       = {{Qwen Team}},
  title        = {{Qwen/Qwen2.5-VL-7B-Instruct}},
  year         = {2025},
  howpublished = {\url{https://huggingface.co/Qwen/Qwen2.5-VL-7B-Instruct}},
  note         = {Accessed: 2026-04-08}
}

@misc{hf_qwen3_vl_235b_a22b_instruct,
  author       = {{Qwen Team}},
  title        = {{Qwen/Qwen3-VL-235B-A22B-Instruct}},
  year         = {2025},
  howpublished = {\url{https://huggingface.co/Qwen/Qwen3-VL-235B-A22B-Instruct}},
  note         = {Accessed: 2026-04-08}
}

@misc{stockvault_adler32,
  author       = {{Pixabay}},
  title        = {{Fruit Mart}},
  year         = {2016},
  howpublished = {\url{https://www.stockvault.net/photo/200223/adler32}},
  note         = {Accessed: 2026-04-08}
}

@misc{wikimedia_family_ride_bicycle_cycle_trailer,
  author       = {Kamyar Adl},
  title        = {{Family Ride bicycle cycle trailer}},
  year         = {2007},
  howpublished = {\url{https://commons.wikimedia.org/wiki/File:Family_Ride_bicycle_cycle_trailer.jpg}},
  note         = {Accessed: 2026-04-08}
}

@misc{freerange_bowls_of_food,
  author       = {Aline Ponce},
  title        = {{A group of bowls of food}},
  howpublished = {\url{https://freerangestock.com/photos/150988/a-group-of-bowls-of-food.html}},
  note         = {Accessed: 2026-04-08}
}

@misc{picryl_highway_construction_crash,
  author       = {{Pixabay}},
  title        = {{A construction site under a bridge with a crane in the background Highway construction site valley bridge crash}},
  year         = {2016},
  howpublished = {\url{https://picryl.com/media/highway-construction-site-valley-bridge-crash-dc08bd}},
  note         = {Accessed: 2026-04-08}
}

@article{lowe2004distinctive,
  title={{Distinctive Image Features from Scale-Invariant Keypoints}},
  author={Lowe, David G},
  journal={{International Journal of Computer Vision}},
  volume={60},
  number={2},
  pages={91--110},
  year={2004},
  publisher={Springer}
}

@inproceedings{lindenberger2023lightglue,
  title={{LightGlue: Local Feature Matching at Light Speed}},
  author={Lindenberger, Philipp and Sarlin, Paul-Edouard and Pollefeys, Marc},
  booktitle={Proceedings of the IEEE/CVF international conference on computer vision},
  pages={17627--17638},
  year={2023}
}

@article{hu2022lora,
  title={{LoRA: Low-Rank Adaptation of Large Language Models}},
  author={Hu, Edward J and Shen, Yelong and Wallis, Phillip and Allen-Zhu, Zeyuan and Li, Yuanzhi and Wang, Shean and Wang, Liang and Chen, Weizhu and others},
  journal={ICLR},
  volume={1},
  number={2},
  pages={3},
  year={2022}
}

@inproceedings{sasagawa2025constructing,
  title={{Constructing Multimodal Datasets from Scratch for Rapid Development of a Japanese Visual Language Model}},
  author={Sasagawa, Keito and Maeda, Koki and Sugiura, Issa and Kurita, Shuhei and Okazaki, Naoaki and Kawahara, Daisuke},
  booktitle={Proceedings of the 2025 Conference of the Nations of the Americas Chapter of the Association for Computational Linguistics: Human Language Technologies (System Demonstrations)},
  pages={470--484},
  year={2025}
}

@article{zheng2023judging,
  title={{Judging LLM-as-a-Judge with MT-Bench and Chatbot Arena}},
  author={Zheng, Lianmin and Chiang, Wei-Lin and Sheng, Ying and Zhuang, Siyuan and Wu, Zhanghao and Zhuang, Yonghao and Lin, Zi and Li, Zhuohan and Li, Dacheng and Xing, Eric and others},
  journal={{Advances in Neural Information Processing Systems}},
  volume={36},
  pages={46595--46623},
  year={2023}
}

@inproceedings{lin2004rouge,
  title={{ROUGE: A Package for Automatic Evaluation of Summaries}},
  author={Lin, Chin-Yew},
  booktitle={Proc. Workshop on Text Summariation Branches Out, Post-Conference Workshop of ACL 2004},
  year={2004}
}
\clearpage

\appendix
\section{Appendix}
\label{sec:appendix}

\subsection{Prompt Example for Factory Task Analysis}
\label{sec:appendix-factory-prompt}

During inference for task analysis, we utilized prompts that provided visual descriptions of tools as hints, leveraging the fact that each task is strongly associated with a particular tool. Providing only tool names can lead to unstable recognition for highly specialized tools. Therefore, we incorporated visual cues into the prompt, such as shape, color, and characteristic components (e.g., ratchet mechanisms, socket sections, and cables), to assist the model in visual identification. The following are representative examples in English and Japanese:

English prompt example:
\begin{adjustwidth}{2em}{0pt}
\color{promptgreen}
\setlength{\parindent}{0pt}
\setlength{\parskip}{0pt}
    \texttt{Identify the task currently being performed by the worker in the image. Select the corresponding one from the following ten task labels.\\
    Because each task is characterized by a particular tool, use the visual description of the tool as a hint.\\
    The task labels and tool characteristics are as follows:\\
    A: A long metal torque wrench. It has a knurled grip, a ratchet mechanism, and a heavy socket head.\\
    B: A soldering iron. It has a yellow resin handle, a cylindrical metal heater section, and a thin silver tip.\\
    \ldots\\
    J: An industrial glue gun. It has a white resin body, a large trigger mechanism, and a thick transparent adhesive stick inserted from the back.\\
    Based on the image, choose the task label and output only the corresponding label name. Do not output any other text.}
\end{adjustwidth}

\vskip\baselineskip
Japanese prompt example:
\begin{adjustwidth}{2em}{0pt}
\color{promptgreen}
\setlength{\parindent}{0pt}
\setlength{\parskip}{0pt}
    \texttt{作業者が画像内で現在行っているタスクを特定してください。以下の10個のタスクラベルから該当するものを選んでください。\\
    各タスクは特定の工具で特徴づけられるため、工具の視覚的な説明をヒントとして用いてください。\\
    タスクラベルと工具の特徴は以下のとおりです：\\
    A: 金属製の長いトルクレンチ。ローレット加工のグリップ、ラチェット機構、重厚なソケットヘッドを持つ。\\
    B: はんだごて。黄色い樹脂製ハンドル、円筒状の金属製ヒータ部、先端が細い銀色のチップを持つ。\\
    \ldots\\
    J: 工業用グルーガン。白い樹脂ボディ、大きなトリガー機構、背面から太い透明な接着スティックが差し込まれている。\\
    画像に基づいてタスクラベルを選び、該当するラベル名のみを出力してください。その他の文章は出力しないでください。}
\end{adjustwidth}
\subsection{Definition of Evaluation Metrics for Anomaly Detection}
\label{sec:appendix-anomaly-metrics}
For anomaly detection evaluation, an F1-score is computed individually for each sample (comprising a target and a reference image pair), and the macro-average across all samples is reported. If all samples were evaluated globally at the bounding box level, images containing numerous anomalous objects would disproportionately influence the final score. By computing an F1-score per sample, we evaluate image-level detection performance while mathematically treating all samples equally.

For the $k$-th sample, let $TP_k$ be the number of correct matches between predictions and ground truth, $FP_k$ be the number of unmatched predictions, and $FN_k$ be the number of unmatched ground-truth instances. The F1-score $F_k$ is given by:
\begin{align}
F_k &= \frac{2TP_k}{2TP_k + FP_k + FN_k}.
\end{align}

For $N$ samples, the average F1-score $M$ is defined as:
\begin{align}
M &= \frac{1}{N}\sum_{k=1}^{N} F_k.
\end{align}

Depending on the operational requirement, the average F1-score is evaluated under two distinct conditions:
\begin{itemize}
    \item \textbf{bbox only} (position only): Predictions satisfying the bounding box matching criterion are counted as $TP_k$.
    \item \textbf{bbox + label} (position + content): Predictions satisfying both the bounding box and label matching criteria are counted as $TP_k$.
\end{itemize}

The matching criteria are defined as follows:
\begin{itemize}
    \item \textbf{bbox match}: The IoU between the predicted bounding box and the ground-truth bounding box is $\ge 0.5$.
    \item \textbf{label match}: The predicted label and the ground-truth label as strings are identical.
\end{itemize}

\subsection{Example Outputs for VQA Tasks}
\label{sec:appendix-vqa-examples}
Below are several examples of VQA outputs using PLaMo 2.1-8B-VL. In the following examples, the original Japanese questions and instructions provided to the model are shown first, with English translations provided in parentheses for readability.

\begin{minipage}{\textwidth}
\noindent\textbf{Example 1 -- VQA}

\begin{center}
    \includegraphics[width=0.78\linewidth]{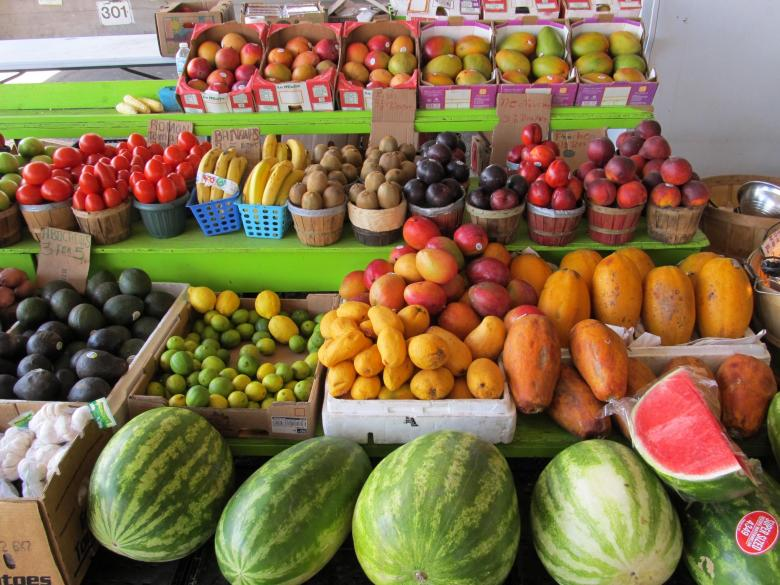}
\end{center}

{\footnotesize
Photo: Pixabay ``Fruit Mart''. Source: Stockvault. License: Creative
Commons - CC0. URL:~\cite{stockvault_adler32} \par
}

\begin{itemize}[leftmargin=1.5em,itemsep=0.2em,topsep=0.2em]
    \item Question: 前方に積まれているのは何ですか？ (What is stacked in the front?)
    \item Answer from PLaMo 2.1-8B-VL: スイカ (Watermelon)
    \item Capability illustrated: The ability to use the relative positional expression ``in the front'' in the image as a cue and identify the object located in the corresponding region (watermelon) (object recognition based on spatial relations)
\end{itemize}
\end{minipage}

\par\medskip

\begin{minipage}{\textwidth}
\noindent\textbf{Example 2 -- VQA}

\begin{center}
    \includegraphics[width=0.78\linewidth]{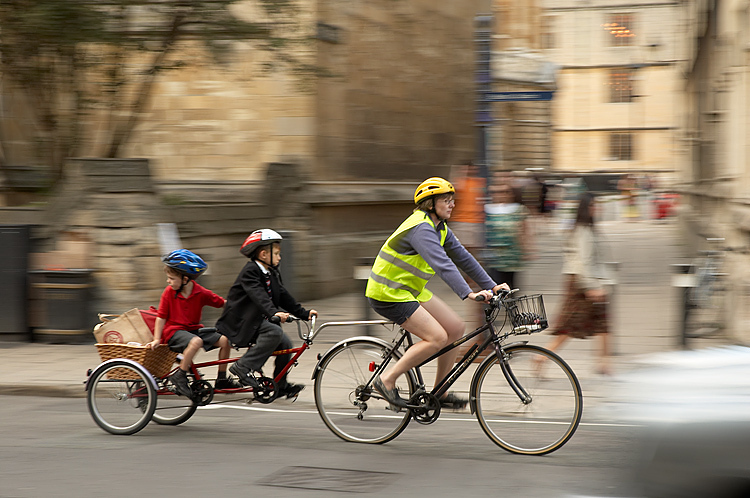}
\end{center}

{\footnotesize
Photo: Kamyar Adl ``Family Ride bicycle cycle trailer''. Source:
Wikimedia Commons (the original image was published on Flickr). License:
Creative Commons Attribution 2.0 Generic (CC BY 2.0,~\cite{cc_by_20}).
URL:~\cite{wikimedia_family_ride_bicycle_cycle_trailer}. \par
}

\begin{itemize}[leftmargin=1.5em,itemsep=0.2em,topsep=0.2em]
    \item Question: 自転車に乗っている人の中で一番小さい人のヘルメットの色は何色ですか？ (Among the people riding bicycles, what color is the helmet of the smallest person?)
    \item Answer from PLaMo 2.1-8B-VL: 青色 (Blue)
    \item Capability illustrated: The ability to distinguish the smallest person among multiple people riding bicycles and answer the attribute of that person's helmet color (comparison and relational reasoning + attribute recognition)
\end{itemize}
\end{minipage}

\par\medskip

\begin{minipage}{\textwidth}
\noindent\textbf{Example 3 -- VQA}

\begin{center}
    \includegraphics[width=0.78\linewidth]{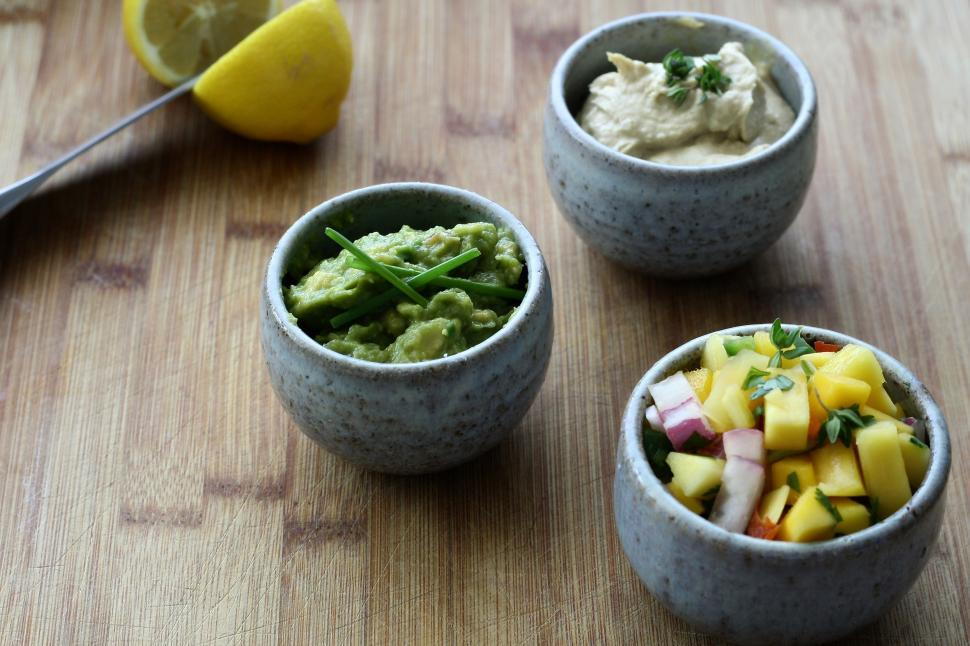}
\end{center}

{\footnotesize
Photo: Aline Ponce ``A group of bowls of food''. Source: Freerange
Stock. License: CC0 1.0 (Creative Commons CC0). URL:~\cite{freerange_bowls_of_food}. \par
}

\begin{itemize}[leftmargin=1.5em,itemsep=0.2em,topsep=0.2em]
    \item Question: 黄色のものが入っていないボウルは全部でいくつありますか？ (How many bowls do not contain anything yellow?)
    \item Answer from PLaMo 2.1-8B-VL: 2つ (2)
    \item Capability illustrated: The ability to distinguish bowls from other objects (such as lemons) and count only the bowls that satisfy the condition ``do not contain anything yellow'' (conditional instance counting)
\end{itemize}
\end{minipage}

\par\bigskip

\subsection{Example Outputs for Visual Grounding Tasks}
\label{sec:appendix-vg-examples}
Below are several examples of Visual Grounding outputs using PLaMo 2.1-8B-VL. For each instruction, the model's predicted bounding box is overlaid on the corresponding region in the image. In the following examples, the original Japanese questions and instructions provided to the model are shown first, with English translations provided in parentheses for readability.

\begin{minipage}{\textwidth}
\noindent\textbf{Example 1 -- VG}

\begin{center}
    \includegraphics[width=0.78\linewidth]{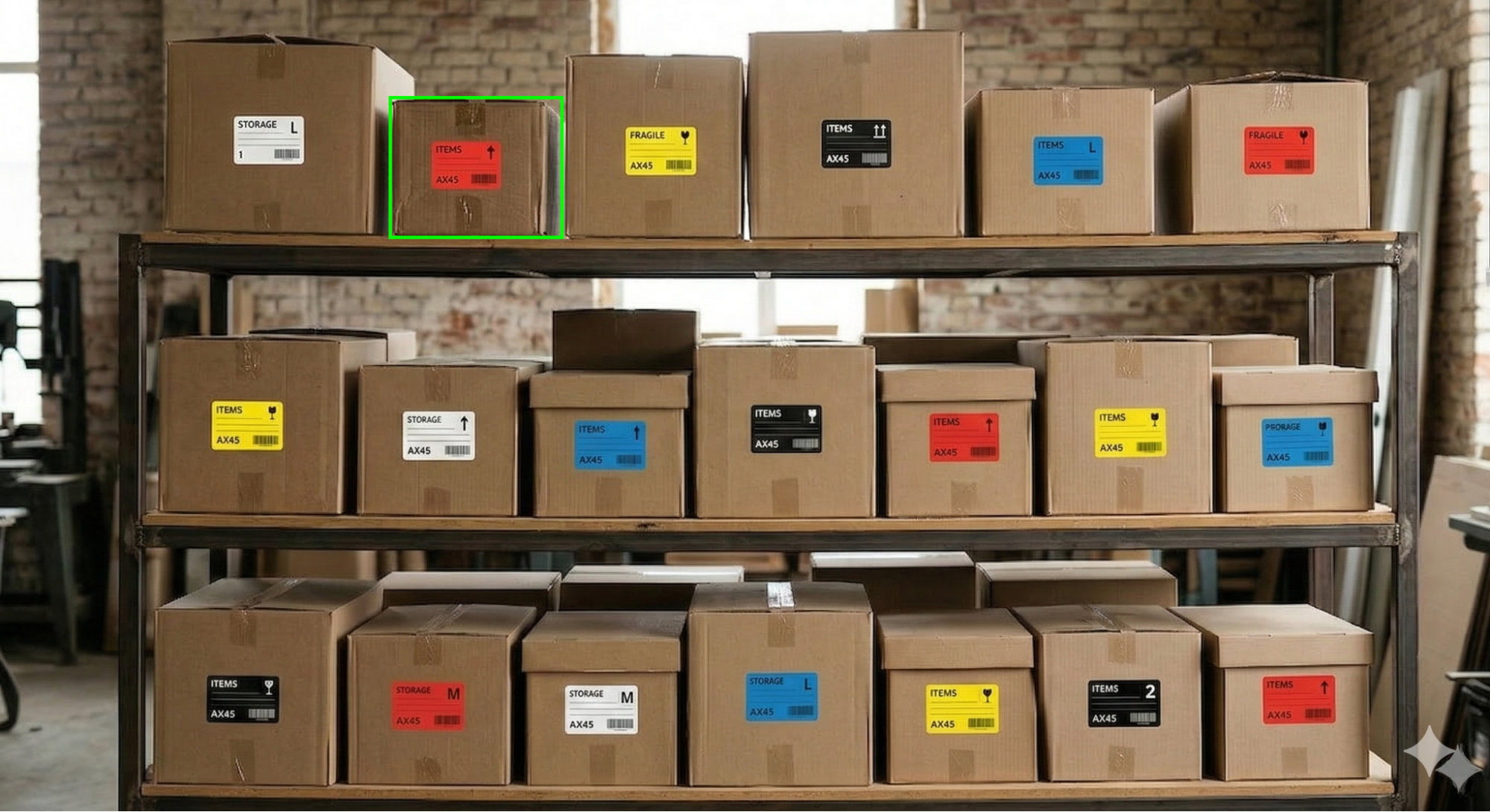}
\end{center}

{\footnotesize
Image generated using Gemini Nano Banana 2. For inclusion in this report, an additional bounding-box overlay has been added.\par
}

\begin{itemize}[leftmargin=1.5em,itemsep=0.2em,topsep=0.2em]
    \item Instruction 1: 上の段にある左側にある赤いタグの段ボールを検出してください。 (Detect the cardboard box with the red tag on the left side of the top shelf.)
    \item Capability illustrated: The ability to uniquely identify the target among multiple cardboard boxes by understanding the composite conditions ``top shelf,'' ``left side,'' and ``red tag,'' and to output its region (understanding positional expressions + reference resolution based on attributes)
\end{itemize}
\end{minipage}

\par\medskip

\begin{minipage}{\textwidth}
\noindent\textbf{Example 2 -- VG}

\begin{center}
    \includegraphics[width=0.78\linewidth]{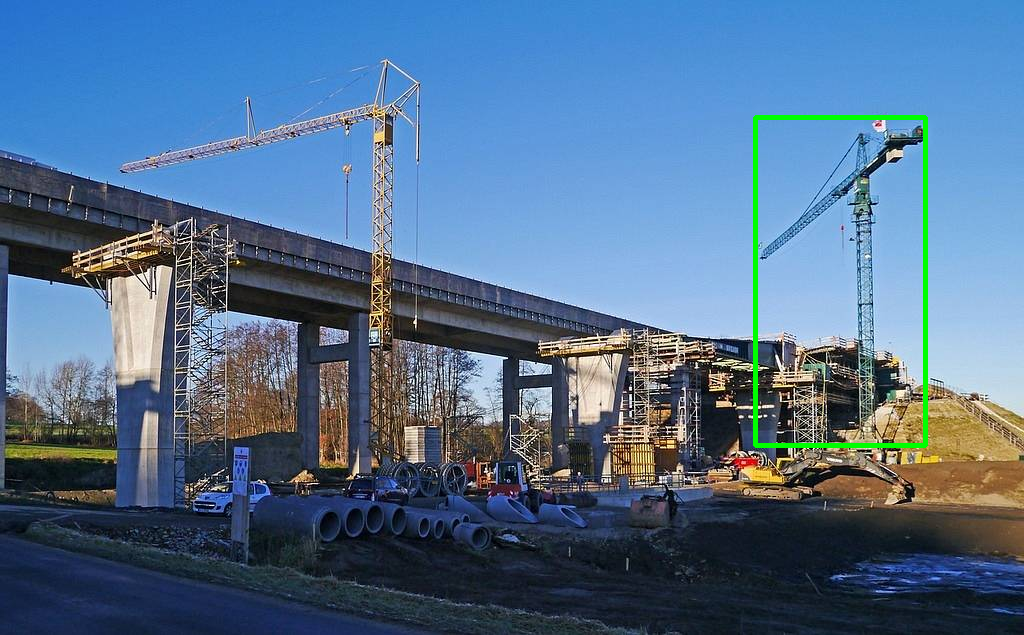}
\end{center}

{\footnotesize
Photo: Pixabay (via PICRYL). License: CC0 1.0 Universal (Public Domain
Dedication). URL:~\cite{picryl_highway_construction_crash}. For
inclusion in this report, an additional bounding-box overlay has been
added.\par
}

\begin{itemize}[leftmargin=1.5em,itemsep=0.2em,topsep=0.2em]
    \item Instruction 2: ショベルカーの近くにあるクレーンを検出して。 (Detect the crane near the excavator.)
    \item Capability illustrated: The ability to identify the target among multiple cranes using the spatial relation expression ``near the excavator'' as a cue and output its region (understanding relational expressions + spatial reasoning)
\end{itemize}
\end{minipage}

\end{document}